\documentclass[10pt,twocolumn,letterpaper]{article}

\usepackage[pagenumbers]{wacv} 

\usepackage{graphicx}
\usepackage{amsmath}
\usepackage{amssymb}
\usepackage{booktabs}
\usepackage{multibib}

\usepackage{times}
\usepackage{epsfig}
\usepackage{color, colortbl}
\usepackage{xcolor}
\usepackage{tikz}
\usetikzlibrary{shapes.geometric,angles,quotes,positioning}
\usetikzlibrary{arrows.meta}
\usepackage{pgfplots}
\usepackage{booktabs}
\usepackage{pgfplotstable}
\usepackage{array,multirow}
\usepackage{balance}
\usepackage{caption}
\usepackage{subcaption}
\usepackage{setspace}
\usepackage{fix-cm}

\pgfplotsset{compat=1.18}

\usepackage{tabularx}
\usepackage{etoolbox,siunitx}

\usepackage{navigator}

\makeatletter
\newcommand*{\addFileDependency}[1]{
  \typeout{(#1)}
  \@addtofilelist{#1}
  \IfFileExists{#1}{}{\typeout{No file #1.}}
}
\makeatother

\definecolor{Lightgray}{gray}{0.9}
\usepackage{pifont}
\newcommand{\xmark}{\ding{55}}%

\definecolor{our_blue}{HTML}{ECD9ED}

\newcommand{\pseudoparagraph}[1]{\paragraph{#1} }
\usepackage{arydshln}

\definecolor{calmgreen}{rgb}{0.2, 0.7, 0.1}
\usepackage[pagebackref=false,breaklinks,citecolor={calmgreen},colorlinks]{hyperref}

\usepackage[capitalize]{cleveref}
\crefname{section}{Sec.}{Secs.}
\Crefname{section}{Section}{Sections}
\Crefname{table}{Table}{Tables}
\crefname{table}{Tab.}{Tabs.}

\newcommand\blfootnote[1]{%
  \begingroup
  \renewcommand\thefootnote{}\footnote{#1}%
  \addtocounter{footnote}{-1}%
  \endgroup
}

\usepackage{fancyhdr}

\newcites{supp}{References}

\begin{document}

\title{Masked Event Modeling: Self-Supervised Pretraining for Event Cameras}

\author{Simon Klenk$^{1,2\,*}$\;
David Bonello$^{1\,*}$\;
Lukas Koestler$^{1,2\,*}$\;
Nikita Araslanov$^{1,2}$\;
Daniel Cremers$^{1,2}$\\
$^1\:$Technical University of Munich \quad $^2\:$Munich Center for Machine Learning\\
{\tt\small \{simon.klenk, david.bonello, lukas.koestler, nikita.araslanov, cremers\}@tum.de } \\
\normalsize{$^*\:$Equal contribution}
}

\maketitle
\thispagestyle{fancy}

\begin{abstract}
   Event cameras asynchronously capture brightness changes with low latency, high temporal resolution, and high dynamic range.
   However, annotation of event data is a costly and laborious process, which limits the use of deep learning methods for classification and other semantic tasks with the event modality.
   To reduce the dependency on labeled event data, we introduce Masked Event Modeling (MEM), a self-supervised framework for events. Our method pretrains a neural network on unlabeled events, which can originate from any event camera recording. Subsequently, the pretrained model is finetuned on a downstream task, leading to a consistent improvement of the task accuracy.
   For example, our method reaches state-of-the-art classification accuracy across three datasets, N-ImageNet, N-Cars, and N-Caltech101, increasing the top-1 accuracy of previous work by significant margins.
   When tested on real-world event data, MEM is even superior to supervised RGB-based pretraining. 
   The models pretrained with MEM are also label-efficient and generalize well to the dense task of semantic image segmentation.\blfootnote{Code and models: \url{https://github.com/tum-vision/mem}.}
\end{abstract}

\begin{figure}
    \centering
    
    \definecolor{blue}{HTML}{0A47C2}
    \definecolor{red}{HTML}{CC0029}
    \definecolor{brown}{HTML}{188B65}
    
    \begin{tikzpicture}
        \begin{axis}[ymax=89, xmax=110, ymin=25, xmin=0, width=0.5 \textwidth, height=8cm, legend pos=south east, ytick={10,30,40,50,60,70,80,90}, xtick={10,20,50,100}, minor xtick={10,20,30,40,50,60,70,80,90,100}, axis x line=bottom, axis y line=left, grid=both, minor x tick num=3, log ticks with fixed point, line width=1pt, every axis plot/.append style={ultra thick}, xlabel = Labeled data{,} \%, ylabel=Top-1 accuracy{,} \%, legend cell align={left}, ylabel near ticks]
            \addplot+ coordinates {(10,66.78) (20,73.58) (50,80.14) (100,85.60)};
            \addplot+[mark=*, dashed, mark options={solid}] coordinates {(10,36.76) (20,42.98) (50,55.34) (100,66.94)};
            \legend{MEM (Ours), ViT-from-scratch}
        \end{axis}
        
    \end{tikzpicture}
    \caption{Classification accuracy on N-Caltech101 \cite{orchard2015NCaltech} with limited labeled data. 
    Our self-supervised pretraining method MEM \textit{(blue)} greatly improves the baseline \textit{(red)} trained from random initialization. The benefits are increasingly pronounced with a scarce amount of labeled data.}
    \label{fig:comparison_pt}
    \vspace{-0.2cm}
\end{figure}
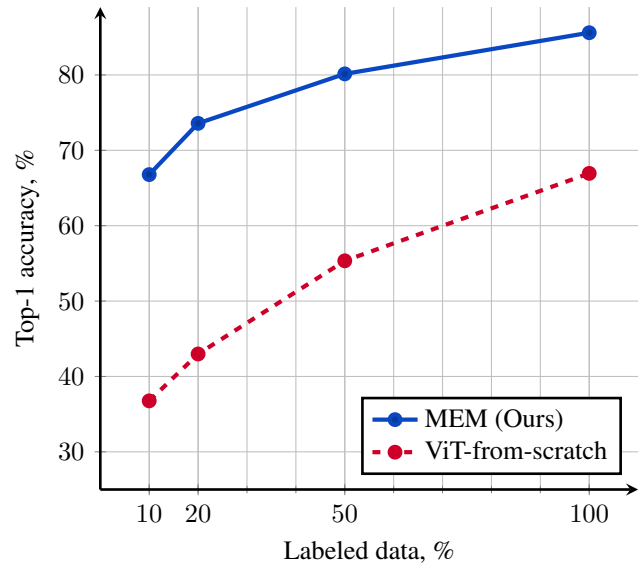

\section{Introduction}
\label{sec:intro}
Event cameras are promising imaging sensors for robotics and virtual reality (VR).
They have the potential to enable applications previously inaccessible for frame-based (RGB) cameras, such as object classification in high-speed autonomous driving or eye and hand tracking in low-power VR systems.
This is because event cameras track pixels independently, triggering them asynchronously only if their brightness change exceeds a threshold \cite{gallego2020Survey}.
Such property allows them to have high temporal resolution, high dynamic range (HDR), low latency, and low power consumption. 

Addressing high-level vision tasks from event data, such as object classification, is significantly more challenging than using RGB images.
RGB-based approaches profited significantly from large labeled training datasets \cite{sun2017revisiting}.
Indeed, the remarkable progress in this field can be largely attributed to the abundance of labeled datasets. 
However, the event-based modality is temporal and lacks color and texture, which impedes the creation of large annotated event-based datasets.
Despite the recently released event-based datasets like N-ImageNet \cite{kim2021nimnet} and N-EPIC-Kitchens \cite{plizzari2022e2go}, \emph{labeled} event data is still in short supply \cite{gallego2020Survey, messikommer2022bridging, sironi2018hats}. 

One solution to counteract the dependency on large-scale labeled datasets is self-supervised learning (SSL).
SSL techniques have already precipitated a significant advance in the natural language processing (NLP) and vision communities \cite{bao2021beit, bardes2021vicreg, caron2021emerging, chen2020simpleContrastive, devlin2018bert, grill2020bootstrap, he2022masked, jing2020self}.
SSL consists of two consecutive stages.
First, a neural network is pretrained without labels by solving a pretext task. For example, BERT~\cite{devlin2018bert} performs pretraining on a large corpus of unlabeled text by predicting the words masked out in the input sequence.
The second stage finetunes the network on a downstream task.
Compared to a randomly initialized network, the pretraining is expected to improve network accuracy and label efficiency on the downstream task, as well as to reduce the number of iterations required for finetuning.

In contrast to previous work leveraging RGB images \cite{caron2021emerging,he2022masked}, we aim at pretraining in the event domain. To this end, we develop Masked Event Modeling (MEM). MEM performs self-supervised pretraining on arbitrary event recordings to alleviate the need for labeled events at a large scale. Our approach is close to the recently proposed frame-based method BEIT \cite{bao2021beit} and is inspired by a multitude of proposed extensions to other data modalities \cite{bachmann2022multimae,tan2021vimpac, wang2022bevt, yu2022point}.

\cref{fig:comparison_pt} illustrates the benefit of our pretraining.
Using the parameter initialization provided by MEM, we consistently outperform all event-based baselines from the literature on object classification. We also show the benefit of MEM for semantic segmentation. In summary, our contributions are:
\begin{itemize}
    \item We present a framework for self-supervised learning on the event modality using a simple event masking approach. During pretraining, our method does not require any labels or access to image data, which makes it applicable to any event recording.
    \item We set a new state-of-the-art top-1 accuracy for event-based object classification on all our datasets. On N-ImageNet, MEM surpasses the previous state-of-the-art by +7.96\%. On N-Cars, it improves by +1.49\% and attains up to +14\% improvement on N-Caltech101.
    \item On the downstream task of semantic segmentation, MEM yields an improvement over training from scratch, especially on rare semantic categories.
    \item We empirically show that the common practice of transferring pretrained weights from RGB data to the event domain \cite{gu2021eventdrop, rebecq2019e2vid} is not always optimal.
    For example, on the N-Cars dataset, where the data originates from real-world recordings, we demonstrate that MEM achieves better accuracy than a \emph{supervised} RGB-based pretraining on ImageNet \cite{deng09imgnet1k}.
\end{itemize}

\begin{figure*}
    \centering
     \begin{tikzpicture}[node distance=2cm,auto,baseline={(current bounding box.north)}, text opacity=1, fill opacity=0.5, inner sep=0pt]
        \definecolor{blue}{HTML}{0A47C2}
        \definecolor{red}{HTML}{CC0029}
        \definecolor{green}{HTML}{198B66}
    
        \node[inner sep=0pt] (img_events) at (0, 0)
            {\includegraphics[width=\textwidth]{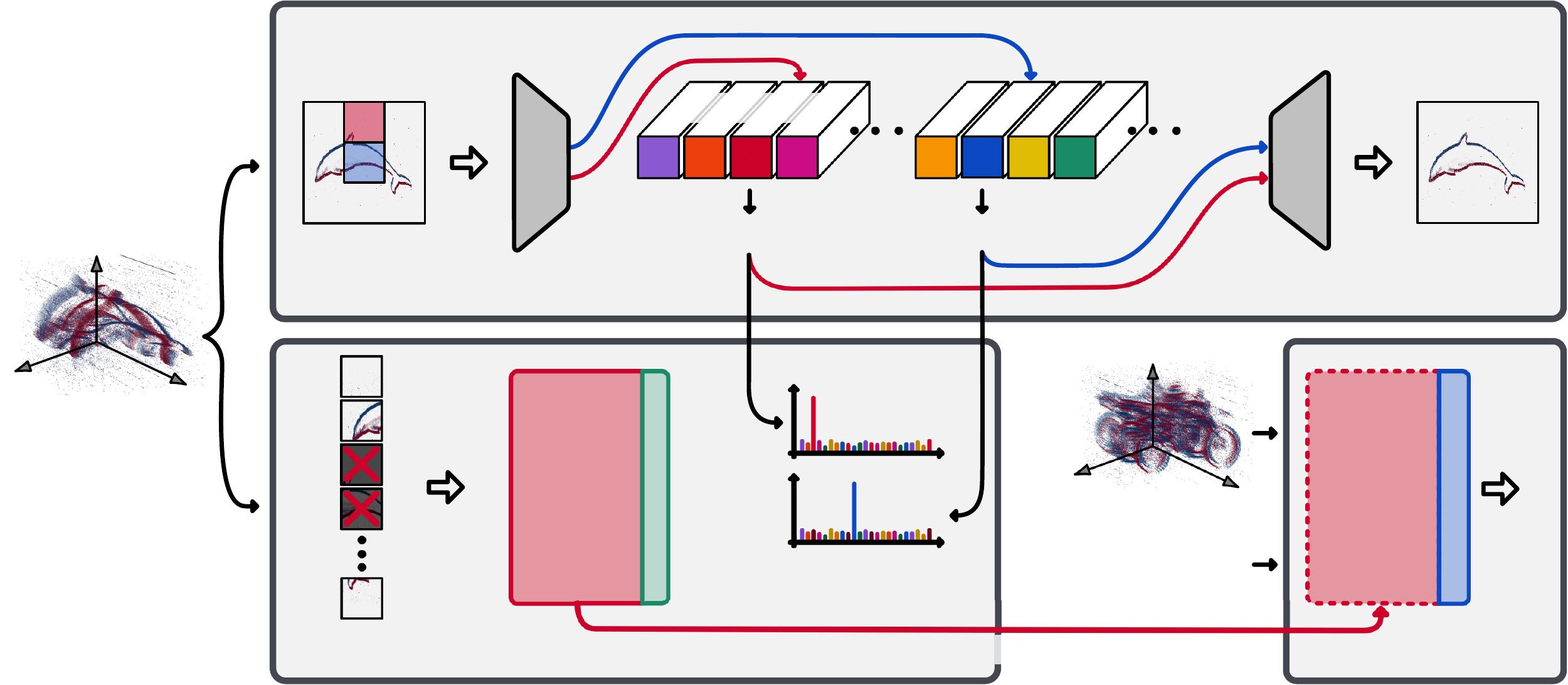}};

            \node[below right] at (-5.4, 3.6) {\large {\color{darkgray} I: dVAE}};
            \node[above right] at (-5.4, -3.7) {\large {\color{darkgray} II: Pretraining}};
            \node[above right] at (5.8, -3.7) {\large {\color{darkgray} III: Finetuning}};

            \node [text width=2cm] at (-7.6, -1) { {\color{darkgray} \begin{center}
                {Unlabeled Events}
            \end{center}}};
            \node [text width=2cm] at (-4.7, 1.1) {\small \color{darkgray} \begin{center}
                 Event Histogram
            \end{center}};
            \node [text width=2cm] at (-2.8, 0.9) {\small \color{darkgray} \begin{center}
                 Event Tokenizer
            \end{center}};
            \node [text width=2cm] at (-0.4, 2.83) {\small \color{darkgray} \begin{center}
                 Codebook
            \end{center}};
            \node [text width=2cm] at (0.95, 1.2) {\small \color{darkgray} Visual Tokens};
            \node [text width=2cm] at (5.75, 0.9) {\small \color{darkgray} \begin{center}
                 Decoder
            \end{center}};
            \node [text width=2cm] at (7.7, 1.3) {\footnotesize \color{darkgray} \begin{center}
                 Reconstruction
            \end{center}};
            \node [text width=3cm] at (0.9, -2.6) {\small \color{darkgray} \begin{center}
                 Predicted masked tokens
            \end{center}};
            \node [text width=3cm] at (4.2, -1.65) {\small \color{darkgray} \begin{center}
                 Labeled Dataset
            \end{center}};
            \node [text width=4cm] at (2.2, -3.23) {\small \color{darkgray} \begin{center}
                 Transfer Weights
            \end{center}};

            \node [] at (-1.4, 2.06) {\small \color{white} 0};
            \node [] at (-0.88, 2.06) {\small \color{white} 1};
            \node [] at (-0.37, 2.06) {\small \color{white} 2};
            \node [] at (0.155, 2.06) {\small \color{white} 3};
            \node [] at (3.09-1.4, 2.06) {\small \color{white} 68};
            \node [] at (3.09-0.88, 2.06) {\small \color{white} 69};
            \node [] at (3.09-0.37, 2.06) {\small \color{white} 70};
            \node [] at (3.09+0.155, 2.06) {\small \color{white} 71};

            \node [] at (-0.38, 1.2) { \color{red} \textbf{2}};
            \node [] at (3.09-0.88, 1.2) { \color{blue} \textbf{69}};
            
            \node [] at (0.34, -1.4) {\scriptsize \color{red} \textbf{2}};
            \node [] at (0.8, -2.4) {\scriptsize \color{blue} \textbf{69}};

            \node [] at (-8.65, -0.37) {\tiny \color{darkgray} x};
            \node [] at (-7.65, 1.08) {\tiny \color{darkgray} y};
            \node [] at (-6.55, -0.5) {\tiny \color{darkgray} t};
            \node [] at (11.8-8.65, -1.15-0.37) {\tiny \color{darkgray} x};
            \node [] at (11.78-7.65, -1.19+1.08) {\tiny \color{darkgray} y};
            \node [] at (11.7-6.55, -1.15-0.5) {\tiny \color{darkgray} t};

            \node [] at (-2.32, -1.6) {\large \color{red} ViT};
            \node [] at (6.55, -1.6) {\large \color{red} ViT};
            \node [rotate=90] at (-1.42, -1.6) {\scriptsize \color{green} Masked Event Modeling};
            \node [rotate=90] at (7.48, -1.6) {\scriptsize \color{blue} Task Layers};
            \node [rotate=90] at (8.37, -1.6) {\large \color{blue} Motorbike};
            \node [text width=1.7cm] at (4.2, -2.3) { \color{blue} \begin{center}
                    + Motorbike
            \end{center}};
            
     \end{tikzpicture}
    
    \caption{An overview of Masked Event Modeling (MEM).
    The proposed method is a three-stage pipeline.
    \textit{(I)}~In the first stage, we train a discrete variational autoencoder (dVAE) \cite{rolfe2016discrete, ramesh2021zeroShotdvae} to compress the input event histograms to a list of discrete visual tokens. Each token -- described by a fixed vector in the codebook -- represents one input patch. The training objective in this stage is event histogram reconstruction.
    \textit{(II)}~In the second stage, we perform self-supervised pretraining of a ViT. The event histogram is divided into patches. We mask 50\% of these patches, and the ViT predicts their corresponding (masked) visual tokens, similar to BERT \cite{devlin2018bert} and BEIT \cite{devlin2018bert}. The masked patches are replaced by a learnable embedding. Since the event tokenizer generates the targets for the ViT, no actual labels are needed. 
    \textit{(III)}~In the final stage, we finetune the pretrained ViT on a downstream task. This is the only stage that requires labeled data. 
}
\vspace{-0.3cm}

\label{fig:pipeline} 
\end{figure*}
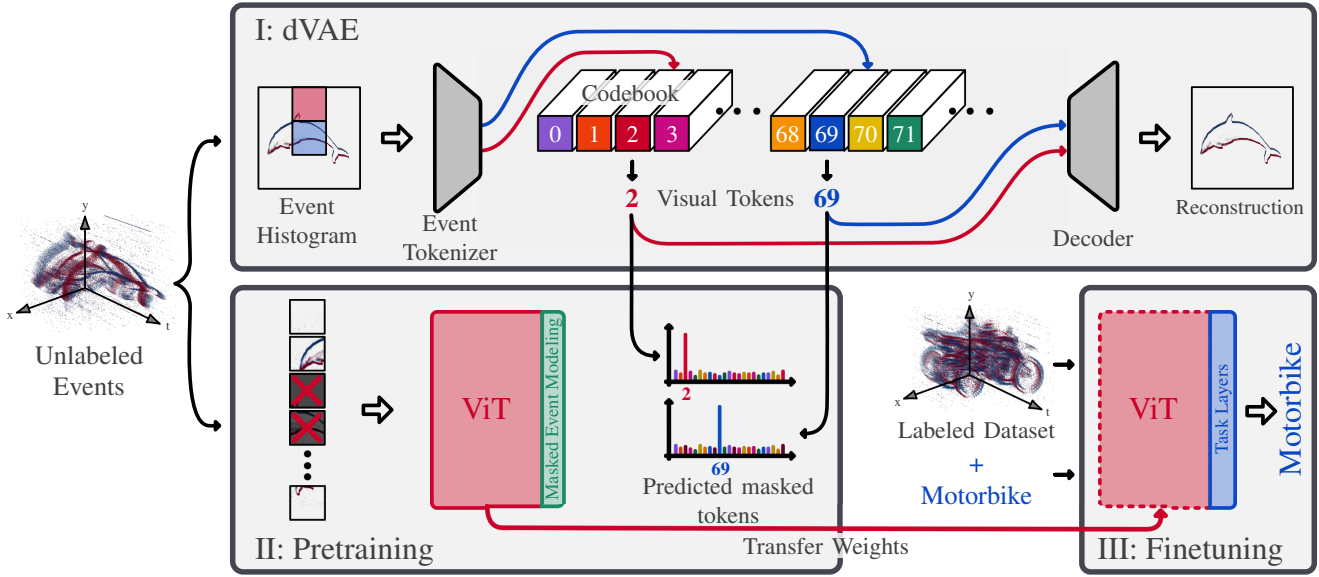

\section{Related Work}
Event-only self-supervised learning is a nascent research avenue.
Therefore, we discuss related work on self-supervised frame-based (RGB) learning and review the methods that tackle the lack of annotated event data.

\pseudoparagraph{Self-supervised learning (SSL)}
The idea of SSL is to first train a network on an unlabeled dataset by solving a pretext task \cite{jing2020self} and then to finetune the network on a downstream task. The pretext task is defined such that the network learns feature representations highly relevant to the downstream task.
After pretraining, the network is finetuned on a downstream task with a small labeled dataset in a supervised fashion. Examples of pretext tasks in vision are image reconstruction from masked or transformed input patches \cite{atito2021sit, he2022masked}, re-ordering of image patches \cite{noroozi2016Jigsaw}, or predicting the degree of image rotations \cite{gidaris2018RotationPred}. 

One notable SSL framework is BEIT \cite{bao2021beit}, which was largely inspired by the recent success of BERT \cite{devlin2018bert} in the NLP domain. Like BERT, the pretext task of BEIT is to mask out sections of an input image with the intention of reconstructing the original content. However, instead of directly predicting the pixels of the masked patches, BEIT predicts visual tokens, which encode semantic information of each patch in a single vector.

While BEIT and related methods \cite{dong2021PECO, fu2021violet, he2022masked, wan2021ICT, xie2022simmim, zhou2021ibot} enhance BERT-style pretraining with the focus on images, several extensions to different data modalities have been recently proposed. Point-BERT \cite{yu2022point} extends the idea to point clouds; BEVT \cite{wang2022bevt} and VIMPAC \cite{tan2021vimpac} apply the reconstruction task for videos; MultiMAE \cite{bachmann2022multimae} jointly reconstructs RGB images with depth and semantic maps.
While Barchid \etal~\cite{barchid2023exploring} take initial steps toward adapting pretraining for events, the accuracy of their method remains limited.

\pseudoparagraph{Overcoming the lack of labeled event data}
Although SSL is not established for event cameras, numerous other works have proposed solutions against the lack of labeled events. For example, Rebecq \etal~\cite{rebecq2019high} show that it is possible to reconstruct grayscale frames from events, which can then be used with conventional frame-based networks. Conversely, events can be derived from existing frame-based datasets with simulation. The drawback of these approaches is the increased computational demand and reconstruction artifacts, which can alter the semantic meaning or neglect event-specific characteristics.

SSL has been used to solve low-level event vision tasks, like optical flow \cite{zhu2018evflo}, intensity reconstruction using a generative event model \cite{paredes2021back}, and, more recently, object classification \cite{yang2023eventPT}. However, these models either work under strong assumptions, such as the availability of temporally synchronized and pixel-aligned grayscale frames or require a substantial amount of RGB data. Semi-supervised approaches also exist \cite{zanardi2019cross, hu2020learning}, but also rely on synchronous recordings of the events and frames.

Recently, a few models have emerged, which leverage unpaired datasets of labeled frames and unlabeled events \cite{messikommer2022bridging, Sun22eccv, wang2021evdistill}. These approaches transfer knowledge from a powerful network in the RGB domain to the event domain. However, these methods depend on labeled images and require datasets of both modalities to be captured under similar conditions, which is not feasible in many applications. 

In contrast to the above methods, we propose a general framework that performs SSL directly on the event data, requiring neither additional labels nor corresponding image data. In most applications, such unlabeled event data is readily available, whereas generating labels or accessing labeled image datasets from the same domain can be very costly or impractical.

\section{Masked Event Modeling}
We propose Masked Event Modeling (MEM), an adaptation of BERT-style pretraining to events \cite{bao2021beit}.
\cref{fig:pipeline} provides an overview. Leveraging the properties of event-based sensors, our method has a conceptual advantage over standard RGB-based models, such as in low-light conditions or applications requiring high temporal resolution.
The events are preprocessed and passed through the MEM pipeline, consisting of three stages.
In the first stage, we train a dVAE to represent the event data in a discrete latent space.
The second pretraining stage employs a self-supervised proxy task of reconstructing partially removed event-based input.
This reconstruction takes place in the latent space learned by the first stage.
In the finetuning stage, we use the weights from pretraining as initialization and finetune the network on a target dataset with labels.

\subsection{Event Processing}\label{ss::pp}
\begin{figure}
    \setlength{\fboxsep}{0pt}%
    \begin{minipage}{0.31\linewidth}
    \fbox{\includegraphics[width=\linewidth]{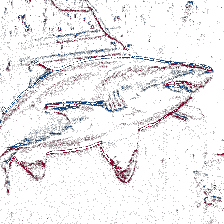}}
    \end{minipage}\hfill
    \begin{minipage}{0.31\linewidth}
    \fbox{\includegraphics[width=\linewidth]{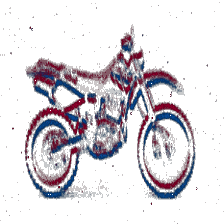}}
    \end{minipage}\hfill
    \begin{minipage}{0.31\linewidth}
    \fbox{\includegraphics[width=\linewidth]{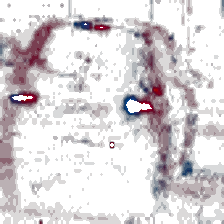}}
    \end{minipage}
    
    \caption{Left to right: Example histograms from N-ImageNet \cite{kim2021nimnet}, N-Caltech101 \cite{orchard2015NCaltech}, and N-Cars \cite{sironi2018hats}. N-Cars is the only real event dataset. It features different event statistics and noise distribution than the two other semi-synthetic dataset (\cf \cref{sec::exps}).}
    \label{fig:samples}
\end{figure}

The raw event data is preprocessed to an event histogram before entering the MEM pipeline.
\cref{fig:samples} illustrates a few examples of event histograms.
Since an event camera asynchronously reports brightness changes at a pixel, the sensor's output is a stream of individual events.
Each event includes a polarity that indicates an increase or decrease in the observed brightness.
For a moving camera under constant illumination, such brightness changes tend to coincide with edges in the conventional image. To obtain an image-like data structure with visible edges, we accumulate the events separated by polarities into a two-channel image $\mathbf{H} \in \mathbb{R}^{H \times W \times 2}$ using up to $N_{\text{max}}$ events. We additionally perform data augmentation on the event histograms. In contrast to related RGB-based works \cite{he2022masked, bao2021beit}, we find that RandAugment \cite{cubuk2020randaugment} is necessary for the best performance with event data (\cf \cref{tab::ablation}).
Note that the input event histogram representation is dense, which enables simple storage, loading, preprocessing, and the use of deeper networks (graph neural networks on sparse representations tend to be relatively shallow \cite{schaefer2022aegnn, li2021graph}).

\subsection{Discrete Variational Autoencoder}\label{ss::VAE}
To reconstruct the masked event histograms during the pretraining phase, we must first reduce the model's output space. Directly predicting raw histogram values of input patches would lead to a higher computational cost and can cause overfitting on low-level visual details \cite{tan2021vimpac}. Instead, we use visual tokens that summarize high-level semantic information in a single vector per patch. Additionally, the tokens are discrete, which allows for formulating the pretraining stage as a classification task.

Following Ramesh \etal~\cite{ramesh2021zeroShotdvae}, we employ a discrete variational autoencoder (dVAE) to learn the compact representation.
The idea is to compress each input patch of size $P_x \times P_y$ to a codebook vector $\textbf{z} \in \mathbb{R}^d$, which summarizes the visual features of the patch. Each codebook vector has a unique index, called a visual token, which later serves as the prediction target in the pretraining phase. Since the codebook is fixed during pretraining, the index of a visual token is sufficient to represent the semantic content of the patch.

The dVAE consists of three parts: the event tokenizer (encoder), the codebook bottleneck $\mathbf{Z} \in \mathbb{R}^{N \times d}$, and the decoder (\cf phase I in \cref{fig:pipeline}). The event tokenizer $q_{\theta}(\textbf{z} \mid \textbf{h})$ takes the full event histogram as the input. Each input patch $\mathbf{h} \in \mathbb{R}^{P_x \times P_y \times 2}$ is mapped to a latent codebook vector $\textbf{z}$. The decoder $p_{\Phi} (\textbf{h} \mid \textbf{z})$ learns to recover the event patch given the visual tokens.
We place a uniform prior on the token distribution.
It can be shown that maximizing the marginal likelihood $p(\mathbf{h})$ corresponds to maximizing the evidence lower bound (ELBO) \cite{kingma2013auto, rezende2014stochastic}:
\begin{multline}
     \mathcal{L}_\text{ELBO}(\theta, \Phi; \mathbf{h}) := \mathop{\mathbb{E}}_{q_{\theta}(\textbf{z} \mid \textbf{h})} [ \log ( p_{\Phi} (\textbf{h} \mid \textbf{z}))]\\
     - D_{KL} \big( q_{\theta}(\textbf{z} \mid \textbf{h}) \, \| \, p_{\Phi} (\textbf{z}) \big).
\label{eq:elbo}
\end{multline}
 
Due to the tokens being discrete, we employ the Gumbel-Softmax relaxation technique \cite{jang2016categorical,maddison2016concrete} for obtaining the gradient, which was previously found useful in related tasks \cite{bao2021beit,ramesh2021zeroShotdvae, yu2022point}. We visualize the decoded tokens in \cref{fig:patches}.
We observe that the tokens capture visual primitives, such as lines and smooth contours. We also found that gradient clipping is required for stable training of the dVAE on event histograms, which is not required in the RGB setting \cite{bao2021beit,he2022masked}.

\subsection{Pretraining}\label{ss::PT}
Our pretraining encourages the network -- in our case, a ViT \cite{dosovitskiy2020image} -- to embed semantic-level content of the event data into the corresponding feature representations.
First, we divide the event histogram into patches and replace 50\%\footnote{We study alternative choices of the masking ratio in \cref{seq::ablation}.} of the patches with learnable mask embeddings, $\mathbf{e}$.
The partially masked input is then used to train the pretraining network, which consists of the ViT and a linear projection (discarded after pretraining).
We denote the set of masked patch indices as $M$.
Using the remaining non-masked patches, $\textbf{h}_{k \notin M}$, and learnable mask embeddings, $\mathbf{e}_{k \in M}$, MEM learns to predict the visual tokens $\textbf{z}_{k \in M}$ of the masked patches. The training objective is 
\begin{equation}
  \text{max} \: \mathbb{E}_{h \sim \mathcal{D}, M \sim \mathcal{M}} \Bigl [ p_{\text{MEM}}(\textbf{z}_{k \in M} \mid \textbf{h}_{k \notin M}, \mathbf{e}_{k \in M}) \Bigr ],
\end{equation}
\noindent where $\mathcal{D}$ is our data distribution; $\mathcal{M}$ models random sampling of patch indices, and $p_{\text{MEM}}$ is the posterior of visual tokens. Following this pretraining process, the ViT can reconstruct masked event histogram, as we illustrate in \cref{fig:recons}.
To succeed on this reconstruction task, the ViT must learn semantically-aware feature representations embedding context information, as we confirm experimentally in \cref{sec::exps}.

\input{tables/fig-recons}

\input{tables/fig-patches}

\subsection{Finetuning}\label{ss::FT}
In the last stage, we transfer the weights of the pretrained network to initialize the ViT for finetuning on a labeled dataset.
A task-specific layer replaces the final linear output layer. The dVAE is no longer used. Since the ViT has learned to complete a partially masked event histogram during pretraining, we expect its weights to effectively process event data for high-level vision tasks. 
Note that our method requires labeled data only for finetuning.

\section{Experiments}\label{sec::exps}
We evaluate the proposed Masked Event Modeling on object classification and semantic segmentation -- the two high-level vision tasks with established benchmarks and well-defined metrics. Recall that event sensors bring advantages over RGB cameras, such as HDR, low power consumption, and high temporal resolution. However, since the events capture only changes in brightness, inferring semantic information from the events only is very challenging.

We report top-1 classification accuracy on N-ImageNet \cite{kim2021nimnet}, N-Caltech101 \cite{orchard2015NCaltech}, and N-Cars \cite{sironi2018hats}. 
We compare MEM on each dataset to three baseline categories: \emph{(i)} baselines from the literature; \emph{(ii)} random weight initialization (ViT-from-scratch), and \emph{(iii)} supervised networks pretrained on RGB data: ImageNet-1k (ViT-1k) and ImageNet-21k (ViT-21k) \cite{deng09imgnet1k,ridnik2121k}. MEM and our baseline methods -- ViT-from-scratch, ViT-1k, and ViT-21k -- use the same implementation, detailed in \cref{sec::hyp}.

\subsection{Object Classification}
\label{subsec::object_classification}

\pseudoparagraph{N-ImageNet}
N-ImageNet \cite{kim2021nimnet} contains 1.78 million event streams recorded by a DVS-Gen3 with resolution 480 × 640 \cite{son2017dvsgen3}. The event camera is moved in front of an LCD monitor, which displays ImageNet-1k \cite{deng09imgnet1k} images containing 1000 classes. It is the largest event-based object classification dataset. The second largest dataset, ASL-DVS \cite{bi2019ASLdvs}, is substantially smaller and features only 24 classes. N-ImageNet is very challenging, as the best-published result so far has only achieved 48.94\% top-1 accuracy, which is significantly below the 90\% mark currently achieved on ImageNet-1k \cite{zhai2022scaling, yu2022coca}.
Pretraining for the N-ImageNet dataset took about 4 days on 4 NVIDIA A100 GPUs and about 2 days on 2 NVIDIA A40 GPUs for the other datasets.

In \cref{tab::nimnet}, we show that our proposed Masked Event Modeling outperforms the baseline N-ImNet-EST \cite{gehrig2019EST} on N-ImageNet by +7.96\%. By contrast, our baseline ViT-from-scratch cannot reach on-par performance with the state of the art. Presumably, this is because training a ViT on ImageNet is challenging due to overfitting \cite{steiner2021HowToVit}.
The results demonstrate that employing MEM pretraining is an effective way to boost performance on this complex event-based classification task.

The shaded top part of \cref{tab::nimnet} shows that our supervised baselines ViT-1k and ViT-21k achieve higher accuracy than our self-supervised MEM. Note that the employed checkpoints for ViT-1k and ViT-21k perform \textit{supervised} pretraining on RGB ImageNet-1k and ImageNet-21k for 300 epochs and require extensive hyperparameter search \cite{steiner2021HowToVit}. As a derivative of ImageNet-1k, N-ImageNet shares many similarities with the original dataset.
Due to computational constraints, we only perform pretraining on N-ImageNet for 75 epochs. We believe longer training and hyperparameter optimization could further reduce the gap between RGB-based and event-only pretraining.

The work by Yang \etal~\cite{yang2023eventPT} is a recent self-supervised pretraining scheme requiring temporally synchronized and aligned pairs of event and RGB frames. Their approach performs comparably to our ViT-21k baseline on N-ImageNet. However, their approach is more restrictive and requires a self-supervised pretrained backbone on ImageNet-21k for extracting a latent representation from the RGB frames.

\bgroup

\begin{table}
\centering

\begin{tabularx}{\linewidth}{X  c@{\hspace{0.5em}}c  S[table-format=2.2]}
\toprule

\multirow{2}{*}{\bfseries Method} & \multicolumn{2}{c}{\bfseries Pretraining} & {\multirow{2}{*}{\bfseries Top-1}} \\
\cmidrule(lr){2-3}
& {Aux. Data\phantom{  }} & {Labels} \\

\midrule
\rowcolor{Lightgray} ViT-1k & ImNet-1k & \checkmark & 61.90 \\
\rowcolor{Lightgray} 
ViT-21k & ImNet-21k & \checkmark & 65.00 \\
\midrule
\rowcolor{Lightgray} 
Yang et al. \cite{yang2023eventPT} & ImNet-21k$^{\dagger}$ & \xmark & 64.83 \\
\midrule
N-ImNet-Hist \cite{maqueda2018event} & \xmark & \xmark & 47.73 \\
N-ImNet-DiST \cite{kim2021nimnet} & \xmark & \xmark & 48.43 \\
N-ImNet-EST \cite{gehrig2019EST} & \xmark & \xmark & \underline{48.93} \\
ViT-from-scratch & \xmark & \xmark & 43.13 \\ 
MEM (ours) & \xmark & \xmark &  \textbf{57.89} \\ 

\bottomrule
\end{tabularx}

\caption{Top-1 classification accuracies on N-ImageNet-1k \cite{kim2021nimnet}. MEM (ours) outperforms all baseline methods which use only event-only. Supervised pretraining on RGB-ImageNet achieves an even higher accuracy. Baseline numbers are from \cite{kim2021nimnet}. \textsuperscript{\textdagger} Yang et al.~\cite{yang2023eventPT} also require paired RGB-event training data, and use a backbone pretrained on ImNet-21k with self-supervision.}
\label{tab::nimnet}
\vspace{-0.3cm}
\end{table}

\egroup 

\pseudoparagraph{N-Caltech101}
Similar to N-ImageNet, N-Caltech101 \cite{orchard2015NCaltech} originates from an event camera moving in front of a monitor displaying images of the original RGB-based Caltech101 \cite{fei2004learning}. It contains 101 classes spread across 8246 samples, each with a timespan of 300 milliseconds.

\bgroup

\begin{table}
\centering

\begin{tabularx}{\linewidth}{X  c  c S[table-format=2.2]}
\toprule
\multirow{2}{*}{\bfseries Method} & \multicolumn{2}{c}{\bfseries Pretraining} & {\multirow{2}{*}{\bfseries Top-1}} \\ 
\cmidrule(lr){2-3}
& {Aux. Data} & {Labels} & \\
\midrule
\rowcolor{Lightgray} 
HATS-Resnet \cite{sironi2018hats} & ImNet-1k & \checkmark & 70.00 \\ 
\rowcolor{Lightgray}
EST \cite{gehrig2019EST} & ImNet-1k & \checkmark & 81.70  \\
\rowcolor{Lightgray}
DVS-ViT \cite{wang2022exploiting} & ImNet-21k & \checkmark & 83.00 \\
\rowcolor{Lightgray} E2VID \cite{rebecq2019e2vid} & ImNet-1k & \checkmark & 86.60  \\
\rowcolor{Lightgray}
EventDrop \cite{gu2021eventdrop} & ImNet-1k & \checkmark & 87.14 \\
\rowcolor{Lightgray}
ACE-BET \cite{liu2022fast} & ImNet-1k & \checkmark & 89.95 \\
\rowcolor{Lightgray} 
ViT-1k & ImNet-1k & \checkmark & 92.06 \\
\rowcolor{Lightgray} 
ViT-21k & ImNet-21k & \checkmark & 91.10 \\

\midrule
HATS \cite{sironi2018hats} & \xmark & \xmark & 64.20 \\
AEGNN \cite{schaefer2022aegnn} & \xmark & \xmark & 66.80 \\
SSRL \cite{barchid2023exploring} & \xmark & \xmark & 72.79 \\
AsynNet \cite{messikommer2020AsyNet} & \xmark & \xmark & 74.50 \\
EvS-S \cite{li2021graph} & \xmark & \xmark & 76.10 \\

ViT-from-scratch & \xmark & \xmark & 66.94 \\
MEM (ours) & \xmark & \xmark & {\underline{85.60}} \\
MEM-NImNet (ours) & NImNet-1k & \xmark & \textbf{90.10} \\

\bottomrule

\end{tabularx}

\caption{Top-1 classification accuracy on N-Caltech101 \cite{orchard2015NCaltech}. MEM (ours) outperforms all baseline methods, which only have access to event information. However, MEM (ours) does not fully close the gap between event-only and RGB-based supervised methods. However, using MEM pretraining with N-ImageNet-1k (MEM-NImNet) outperforms them as well.}
\vspace{-0.3cm}
\label{tab::ncaltech101}
\end{table}

\egroup 
In \cref{tab::ncaltech101}, we show that our method outperforms all baselines that have access only to the labeled events of N-Caltech101 \cite{orchard2015NCaltech} and improves over the previous state of the art by +9.5\%. Remarkably, we can further improve this result by leveraging \textit{unlabeled} event data from N-ImageNet (instead of N-Caltech101) in the pretraining stage, which yields top-1 accuracy of $90.1\%$ (MEM\nobreakdash-NImNet). This nearly closes the gap to fully supervised methods pretrained on the RGB data (\cf shaded rows in \cref{tab::ncaltech101}). MEM\nobreakdash-NImNet stems from pretraining MEM on the N-ImageNet \cite{kim2021nimnet} dataset for 75 epochs and finetuning MEM using the labeled event data from the N-Caltech101 training set.
The fact that we can leverage event data from a different dataset suggests a strong cross-domain generalization of MEM.
Note that MEM\nobreakdash-NImNet requires labels only from N-Caltech101 and relies solely on the event modality.

MEM\nobreakdash-NImNet is only marginally outperformed by our supervised baseline methods, ViT-1k and ViT-21k, which perform slightly better than other methods supervised on the RGB-based ImageNet. This could be explained by our use of advanced data augmentation in the supervised baselines, such as RandAugment \cite{cubuk2020randaugment} (playing an important role, see \cref{seq::ablation}), a cosine learning rate scheduler, and regularization, inspired by state-of-the-art image-based ViT training practice \cite{steiner2021HowToVit,maqueda2018event}, see \cref{sec::hyp} for further details.

\pseudoparagraph{N-Cars}

We also benchmark MEM on the non-synthetic N-Cars dataset \cite{sironi2018hats}, which captures the real-world event dynamics. It was recorded by an event camera mounted behind the windshield of a car in an urban environment. The N-Cars dataset contains 12,336 samples of the class ''Car`` and 11,693 samples of the class ``Background'', where each sample is 100 milliseconds long. 
One fundamental difference between N-Cars and the other datasets is its different spatio-temporal event distributions since the events now originate from a dynamic \emph{3D} scene. By contrast, the event data in N-ImageNet and N-Caltech101 comes from a homography \wrt a flat LCD monitor under constant light \cite{hartley2003multiple}. In contrast, events in N-Cars are also triggered due to brightness changes in the physical world, such as those caused by the blinking of a car's rear lights and traffic lights. As visualized in \cref{fig:samples}, such fluctuations are distinctive marks of the real-world capturing conditions of N-Cars.

\bgroup

\begin{table}
\centering

\begin{tabularx}{\linewidth}{X c@{\hspace{0.5em}}c@{\hspace{1em}}
S[table-format=2.2]}
\toprule
\multirow{2}{*}{\bfseries Method} & \multicolumn{2}{c}{\bfseries Pretraining} & {\multirow{2}{*}{\bfseries Top-1}} \\
\cmidrule(lr){2-3}
& {Aux. Data} & {Labels} & \\

\midrule
\rowcolor{Lightgray} 
HATS-Resnet \cite{sironi2018hats} & ImNet-1k & \checkmark & 90.40 \\
\rowcolor{Lightgray} 
E2VID \cite{rebecq2019e2vid} & ImNet-1k & \checkmark & 91.00  \\
\rowcolor{Lightgray} 
EST \cite{gehrig2019EST} & ImNet-1k & \checkmark & 92.50 \\
\rowcolor{Lightgray} 
EventDrop \cite{gu2021eventdrop} & ImNet-1k & \checkmark & 95.50 \\
\rowcolor{Lightgray} 
ACE-BET \cite{liu2022fast} & ImNet-1k & \checkmark & 97.06 \\
\rowcolor{Lightgray} 
ViT-1k & ImNet-1k & \checkmark & 98.00 \\
\rowcolor{Lightgray} 
ViT-21k & ImNet-21k & \checkmark & 96.24 \\

\midrule
SSRL \cite{barchid2023exploring} & ASL-DVS \cite{yin2020graph} & \xmark  & \underline{95.64} \\
HATS \cite{sironi2018hats} & \xmark & \xmark &  90.20 \\
AsynNet \cite{messikommer2020AsyNet} & \xmark & \xmark & 94.40 \\
EvS-S \cite{li2021graph} & \xmark & \xmark  & 93.10 \\
AEGNN \cite{schaefer2022aegnn} & \xmark & \xmark  & {94.50} \\

ViT-from-scratch  & \xmark & \xmark & 92.71 \\ 
MEM (ours) & \xmark & \xmark & \textbf{98.55} \\
MEM-NImNet (ours) & NImNet-1k\;\; & \xmark & 93.27  \\ 

\bottomrule
\end{tabularx}

\caption{Top-1 classification accuracy on N-Cars \cite{sironi2018hats}. MEM (ours) achieves a new state-of-the-art on this benchmark compared to all baselines. Since N-Cars contains real event data and only two classes, pretraining on ImageNet \cite{deng09imgnet1k} or on N-ImageNet \cite{kim2021nimnet} does not perform as well as for this dataset. Our method requires much less data and compute since it only uses events from N-Cars.}
\label{tab::ncars}
\vspace{-0.3cm}
\end{table}

\egroup

In \cref{tab::ncars}, we show that MEM pretrained only on N-Cars outperforms all presented baselines, including the methods with access to RGB-based pretraining on ImageNet-1k and ImageNet-21k. Compared to ACE-BET \cite{liu2022fast}, we raise the state-of-the-art from 97.06\% to 98.55\% top-1 accuracy, which halves the classification error. While supervised RGB-based pretraining on ImageNet does boost performance on semi-synthetic N-ImageNet (\cf \cref{tab::nimnet}) and N-Caltech101 (\cf \cref{tab::ncaltech101}), this effect is much weaker on the real-world event data in N-Cars, as can be seen by the similar accuracy of the upper and lower part of \cref{tab::ncars}. We hypothesize the reason to be the \emph{domain gap}, \ie the difference in spatio-temporal event statistics of N-Cars compared to the semi-artificial N-ImageNet and N-Caltech101 datasets. This hypothesis can be further supported by our experiment ``MEM\nobreakdash-NImNet'' in the last row of \cref{tab::ncars}, where instead of N-Cars, we use N-ImageNet-1k for pretraining. MEM\nobreakdash-NImNet on N-Cars does not improve significantly over ViT-from-scratch, even though it can access much more event data during pretraining. Recall that, by contrast, pretraining on N-ImageNet -- MEM\nobreakdash-NImNet in \cref{tab::ncaltech101} -- achieves a remarkable finetuning performance on N-Caltech101. This discrepancy strongly suggests that RGB-based pretraining may be limited when the ultimate domain of operation is the event data.
Since MEM operates directly on the event modality, there is no domain gap \wrt the test-time setting, which explains its advantage over RGB-based approaches. 

Overall, the experiments on N-Cars suggest that the common practice in the community of simply transferring RGB-pretrained weights to the event domain is not always the best option. This is especially true if the target task exhibits a specific event distribution distinct from the RGB data used during pretraining. 

\subsection{Semantic Segmentation}\label{sec::ss}
\begin{figure*}
\begin{minipage}[b]{0.35\textwidth}
    \begin{tabularx}{\linewidth}{X
    *{3}{S[table-format=2.2, table-column-width=1.8em]}}
        \toprule
        {\bfseries Method} &  {\bfseries aAcc} & {\bfseries mAcc} & {\bfseries mIoU}  \\
        \midrule
        ViT-from-scratch &  85.07 & 49.90 & 42.79  \\
        MEM (ours) &  \textbf{86.62} & \textbf{51.39} & \textbf{44.62}  \\
        \bottomrule
    \end{tabularx}
    \captionof{table}{Finetuning on DSEC-Semantic \cite{Gehrig21ral} validation: mean IoU (mIoU), all pixel accuracy (aAcc), mean accuracy of each class (mAcc), after 6k steps with batch size 32.}
    \label{tab::dsec}
\end{minipage}\hfill
\begin{minipage}[b]{0.6\textwidth}
    
    \begin{subfigure}[b]{0.3\textwidth}
        \centering
        \begin{picture}(100,100)
        \put(0,0){\includegraphics[width=\textwidth]{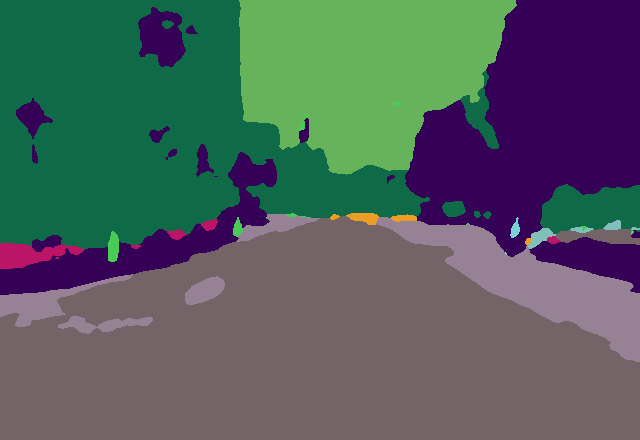}}
        \put(42,7){\footnotesize \textcolor{white}{(a)}}
        \end{picture}
     \end{subfigure}
    \hfill
    \begin{subfigure}[b]{0.3\textwidth}
         \centering
         \begin{picture}(100,100)
         \put(0,0){\includegraphics[width=\textwidth]{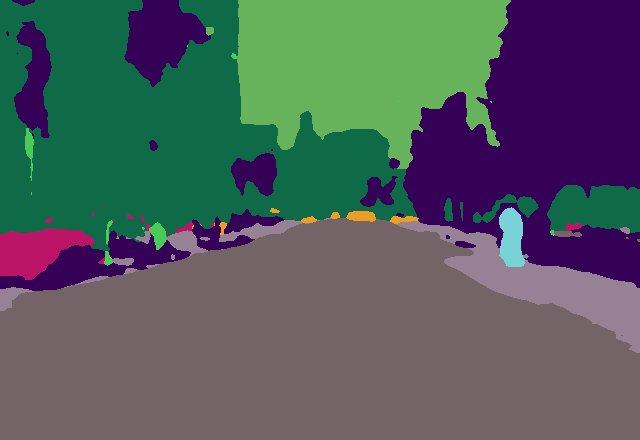}}
         \put(42,7){\footnotesize \textcolor{white}{(b)}}
         \end{picture}
    \end{subfigure}
    \hfill
    \begin{subfigure}[b]{0.3\textwidth}
         \centering
         \begin{picture}(100,100)
         \put(0,0){\includegraphics[width=\textwidth]{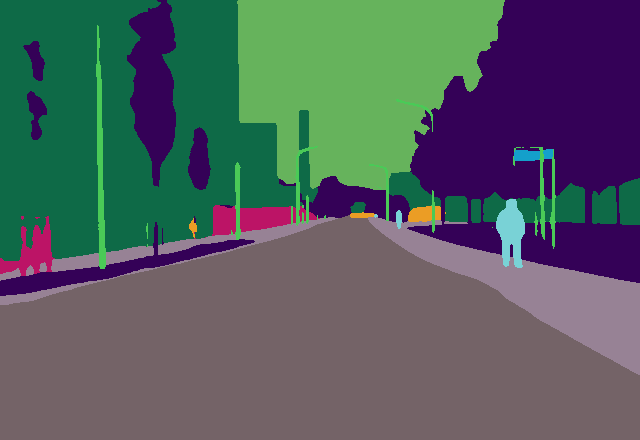}}
         \put(42,7){\footnotesize \textcolor{white}{(c)}}
         \end{picture}
    \end{subfigure}
    
    \captionof{figure}{Semantic segmentation on DSEC-Semantic \cite{Gehrig21ral}. (a)~ViT-from-scratch fails to detect the pedestrian and produces some artifacts on the road. (b)~Our MEM rectifies both issues. (c)~Ground truth. See \cref{supp-sec::supp-segmentation} for more examples.}
    \label{fig:SS}
\end{minipage}
\end{figure*}

As a second downstream task, we evaluate MEM on semantic segmentation -- a high-level vision task of assigning a class label to each image pixel. We use the DSEC-Semantic \cite{Gehrig21ral} dataset.
It contains 10891 event streams annotated with semantic masks. When providing the event histograms to ViT as the input, we resize the original spatial resolution of $640 \times 480$ to $512 \times 512$ to preserve the aspect ratio used by the ViT during pretraining. \cref{tab::dsec} reports quantitative results.
They strongly support our conclusions from the image classification task: the models pretrained with MEM provide a clear advantage over random initialization.
From qualitative results, exemplified by \cref{fig:SS}, we further observe that MEM pretraining improves segmentation accuracy of less dominant semantic categories, such as pedestrians, which can be critical in practice.
\cref{supp-sec::supp-segmentation} provides further examples.

\subsection{Masked Event Modeling with Few Labels}
To demonstrate the benefits of our method in a few-shot scenario, we train MEM with increasingly smaller subsets of the dataset. We split the original train set into mutually exclusive smaller subsets: 100\%, 50\%, 20\%, and 10\%.

MEM is pretrained on the full N-Caltech101 
\cite{orchard2015NCaltech} dataset \textit{without labels}. Afterward, we finetuned the pretrained model on the smaller subsets. MEM is compared with the ViT-from-scratch baseline. Both models have access to the same data and labels during finetuning. In \cref{fig:comparison_pt}, we show that MEM consistently outperforms the baseline ViT-from-scratch. The benefit of MEM becomes increasingly pronounced as the amount of labeled data is reduced. For example, using only 10\% of labeled data -- 650 samples over 101 classes -- MEM achieves 66.78\% top-1 accuracy, whereas ViT-from-scratch reaches only 36.67\%.

\subsection{Ablation Study MEM components}\label{seq::ablation}
\bgroup

\begin{table}
\centering

\begin{tabularx}{\linewidth}{X 
S[table-format=2.2]
S[table-format=2.2]}
\toprule
{\bfseries Ablation Setup} & {\bfseries N-Caltech101}  & {\bfseries N-Cars} \\ 
\midrule
\rowcolor{Lightgray}  Full method & \textbf{85.60} & \textbf{98.55} \\
\midrule
8x8 patches & 80.81 & 96.13 \\
32x32 patches & 78.58 & 97.14 \\
\midrule
25\% mask ratio & 81.16 & 98.47 \\
75\% mask ratio & 82.31 & 97.55 \\
\midrule
33\% pretrain steps & 81.17 & 95.16 \\
\midrule
No randAug \cite{cubuk2020randaugment} & 80.67 & 98.14 \\
\bottomrule
\end{tabularx}

\caption{Ablation study of our method on N-Caltech101~\cite{orchard2015NCaltech} and N-Cars~\cite{sironi2018hats}. Default values for the full method are: patch size 16, masking ratio 50\% and using RandAugment \cite{cubuk2020randaugment}. Event histograms are $224 \times 224$.}
\label{tab::ablation}
\vspace{-0.3cm}
\end{table}

\egroup

We study our method by modifying individual components and report the resulting top-1 accuracy on N-Cars \cite{sironi2018hats} and N-Caltech101 \cite{orchard2015NCaltech}. All other hyperparameters are fixed.
In the first row of \cref{tab::ablation}, we show the base accuracy of MEM on N-Cars or N-Caltech101. We ablate the patch size by changing it to $8 \times 8$ and $32 \times 32$, respectively. The patch size influences the number of visual tokens and the number of patches. We find that the patch size of $16 \times 16$ yields the overall best result. We suspect that if the patch size is small, the information stored inside a patch is not enough to be semantically meaningful. If the patch size is too large, too many concepts must be represented by one patch, negatively influencing the pretraining result.

We next experiment with the masking ratio.
Our default choice of 50\% is superior to  25\% or 75\%.
By comparison, 40\% is the default in BEIT \cite{bao2021beit}; MAE \cite{he2022masked} masks out 75\% of the input patches.
We suspect that a small masking ratio makes the pretraining task too easy, and the ViT tends to learn only a few semantic concepts as a result. Conversely, a high masking ratio makes it hard for the learning to extract semantically relevant features.

We test a shorter schedule of MEM pretraining by only performing 33\% of the pretraining steps. The significant drop in classification accuracy on both datasets suggests that a longer pretraining is crucial. 

Disabling RandAugment \cite{cubuk2020randaugment} results in a performance drop. Our empirical observations on N-ImageNet confirm that RandAugment is effective for event histograms, too. Note that RandAugment is not used in related RGB methods \cite{bao2021beit,he2022masked} (\cf \cref{ss::pp}). This finding and the high downstream accuracy of our baselines ViT-1k and ViT-21k show that modern augmentation techniques are highly beneficial for event data as well and tend to outperform almost all event-specific methods from the literature (\cf \cref{tab::nimnet,tab::ncaltech101,tab::ncars}).

\subsection{Ablation of Pretraining Task}\label{seq::mae}
To validate the choice of our pretraining objective, we also investigate the alternative formulation based on the Masked Autoencoder (MAE) \cite{he2022masked}.
MAE directly reconstructs the masked input instead of indirectly through visual tokens, as we do. 
\cref{tab:mae} reports the results.
We found our MEM reconstruction and the top-1 accuracy to be superior. In order to adapt MAE to events, dubbed as ``eMAE'', we found that the MAE loss needs to be applied to the entire histogram (eMAE-entire-hist) and not just on the masked patches (eMAE-naive-only-mask) as proposed in the original MAE formulation. Nevertheless, it is an interesting insight that MAE can be applied to events with our modification.
Since MAE does not rely on the visual tokens produced by dVAE, the overall framework becomes simpler and hence deserves further investigation.

\begin{table}
    \centering
    \begin{tabularx}{\linewidth}{X S[table-format=2.2]  S[table-format=2.2]}
        \toprule
        {\bfseries Method} & {\bfseries N-Caltech} & {\bfseries N-Cars} \\
        \midrule
    
        (i)\phantom{ii}~MEM (ours) & \textbf{85.60} & \textbf{98.55} \\
        (ii)\phantom{i}~eMAE-naive-only-mask & 58.11 & 95.70 \\
        (iii)~eMAE-entire-hist & 80.61 & 97.50 \\

        \bottomrule
    \end{tabularx}
    \caption{Top-1 accuracy. \emph{(i)} MEM, \emph{(ii)} naive eMAE with loss applied only on masked patches (as proposed by MAE \cite{he2022masked} paper), \emph{(iii)} eMAE-entire-hist loss on the entire histogram (our proposed modification for event-MAE).}
    \label{tab:mae}
    \vspace{-0.3cm}
\end{table}

\subsection{Ablation of Input Representation} \label{sec::input_representation}
Event histograms are our input representation of choice due to their simplicity and practical considerations, such as easy storage and preprocessing.
To justify this design choice, we tested an 8-time-slice input, similar to voxel grids \cite{rebecq2019e2vid}. We did not observe any significant difference in terms of the top-1 accuracy of object classification compared to our representation, as detailed in \cref{supp-sec::input_representation_supp}. Furthermore, our preliminary experiments did not suggest any benefit for object classification if the event histograms encode the event timestamps in a third input channel.
A 2-channel event histogram appears to be sufficient.

\section{Conclusion}
In practice, large amounts of unlabeled events from the target domain are easily accessible, whereas labeling event data is a laborious and costly process. In fact, labeling events is even more difficult than images because of their temporal nature and spatial sparsity (\cf \cref{subsec::object_classification}).
We introduced Masked Event Modeling, MEM, which brings self-supervised learning to event data.
MEM consistently improves the state-of-the-art top-1 classification accuracy across all benchmarks we have tested: by +7.96\% on N-ImageNet, by +1.49\% on N-Cars, and by up to +14\% N-Caltech101.
Furthermore, MEM can also be employed for semantic segmentation with clear accuracy benefits.

Our work also revealed that the common practice of applying an RGB-pretrained network to an event vision task is not always optimal due to the domain gap, \ie when the finetuning dataset has different characteristics from the training data. MEM can address this problem by pretraining on the same type of data modality as the target dataset.

\pseudoparagraph{Limitations} 
While the computational expense of SSL may be amortized over a number of downstream tasks, SSL pretraining generally remains computationally demanding.
Furthermore, despite empirical justification of our input representation, our current framework does not fully leverage the sparsity and high temporal resolution of event cameras.
We have also shown promising results only on two downstream tasks, object classification, and semantic segmentation, but other downstream tasks, such as object detection, may be an interesting testbed as well.
In future work, we plan to explore pretraining formulations for dynamic tasks, such as optical flow estimation.
We will consider using event-specific network architectures, such as asynchronous, sparse (graph) neural networks.
It may also be worthwhile to explore pretraining on neuromorphic hardware, where network training is still challenging.

\pseudoparagraph{Acknowledgements} This work was supported by the ERC Advanced Grant SIMULACRON.

\balance
{\small
\bibliographystyle{ieee_fullname}
\bibliography{egbib}
}

\clearpage
\pagenumbering{roman}
\appendix

\twocolumn[
  \begin{@twocolumnfalse} 
  {
    \centering
    \Large \textbf{Masked Event Modeling: Self-Supervised Pretraining for Event Cameras}\\[0.5em]
    \large \textbf{-- Supplementary Material --}\\[2em] 
  }
  \end{@twocolumnfalse}
]

\section{Semantic Segmentation: More Examples}
\label{supp-sec::supp-segmentation}

\begin{figure*}[h]
    \begin{tabular}{c c c}
    \includegraphics[width=0.31\textwidth]{figs/fig-SS/000346_fromScratch.png}
    & \includegraphics[width=0.31\textwidth]{figs/fig-SS/000346_MEM.png}
    & \includegraphics[width=0.31\textwidth]{figs/fig-SS/000346_GT.png}
    \\
    \includegraphics[width=0.31\textwidth]{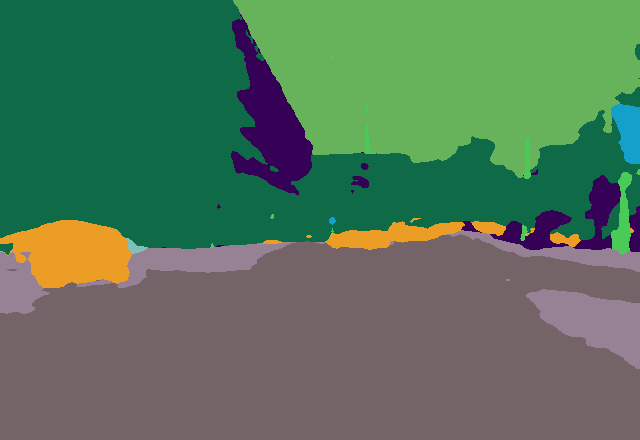}
    & \includegraphics[width=0.31\textwidth]{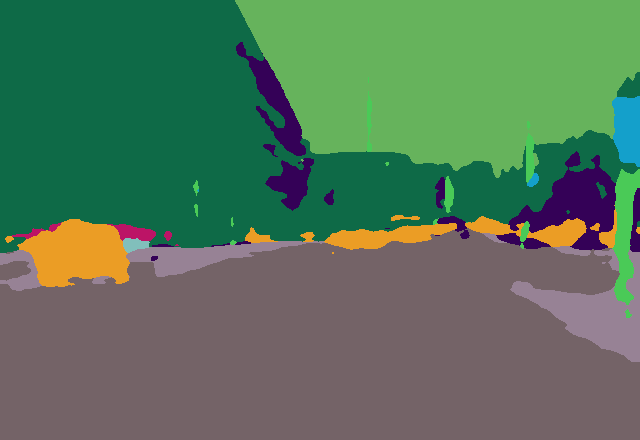}
    & \includegraphics[width=0.31\textwidth]{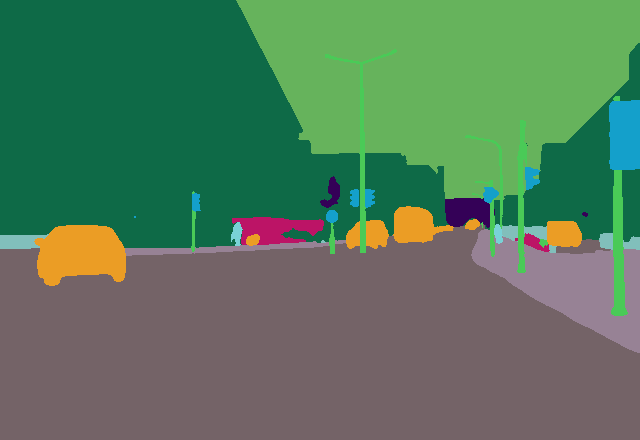}
    \\
    \includegraphics[width=0.31\textwidth]{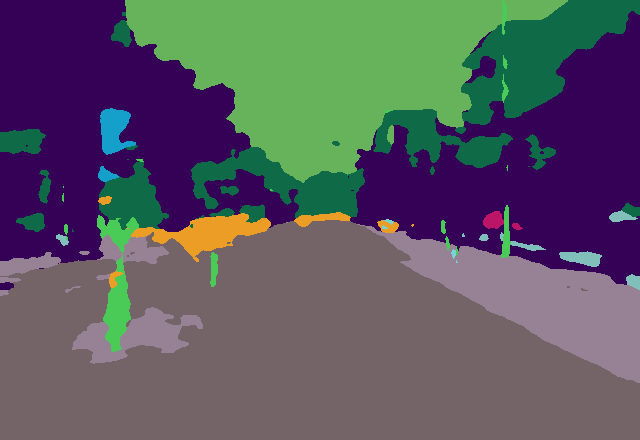}
    & \includegraphics[width=0.31\textwidth]{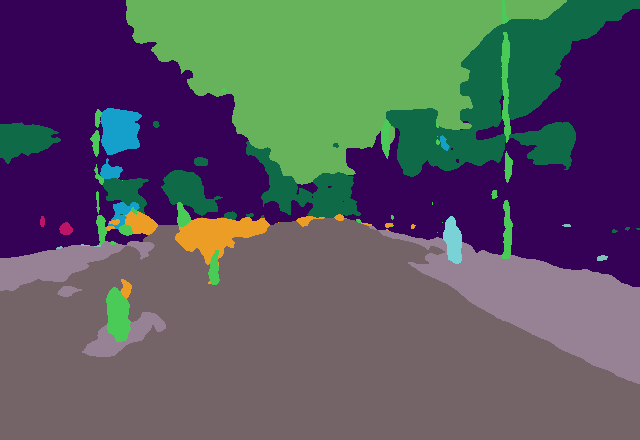}
    & \includegraphics[width=0.31\textwidth]{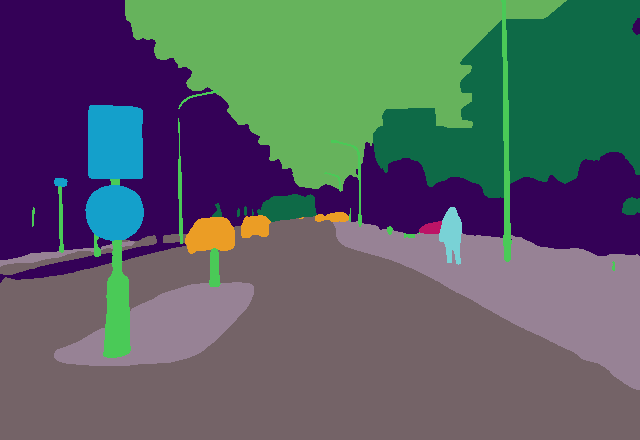}
    \\
    \includegraphics[width=0.31\textwidth]{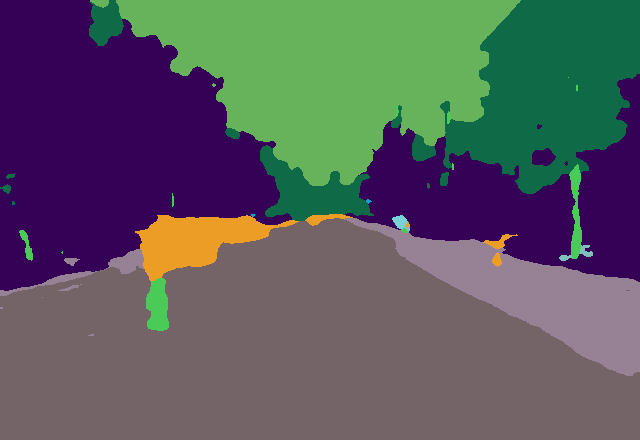}
    & \includegraphics[width=0.31\textwidth]{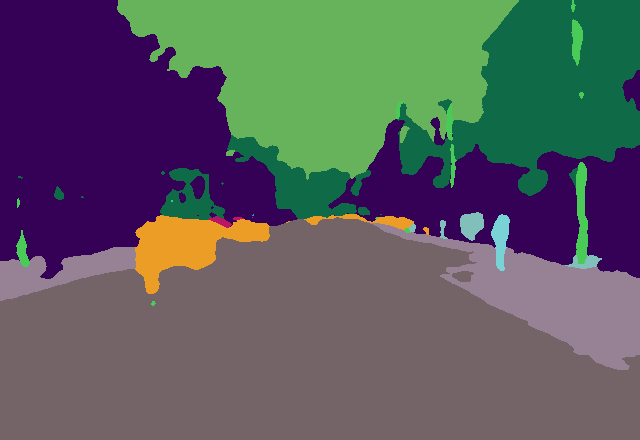}
    & \includegraphics[width=0.31\textwidth]{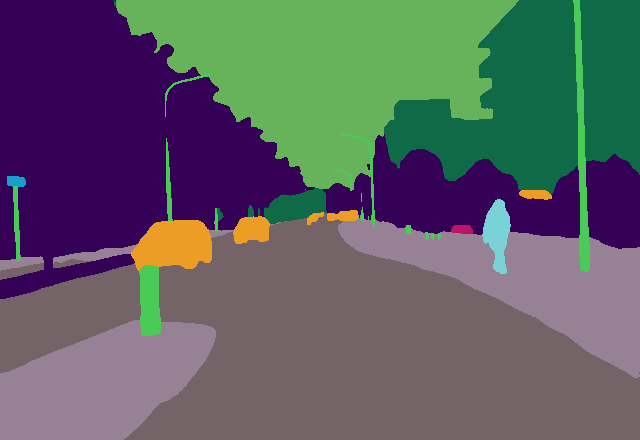}
    \\
    (a) ViT-from scratch & (b) MEM (ours)& (c) Ground truth \\
    \end{tabular}
\caption{Semantic segmentation examples: (a)~ViT-from-scratch, (b)~MEM (ours), (c) the ground truth. MEM recovers the pedestrians (the right image half, 1st, 3rd, and 4th row), as well as the lamp pole (the right image half, 2nd, 3rd, and 4th row) more reliably.}
\vspace{-0.5em}
\label{fig:ssEx}
\end{figure*}

\begin{figure*}[h]
	\centering
    \begin{tabular}{c c c}
    \includegraphics[width=0.31\linewidth]{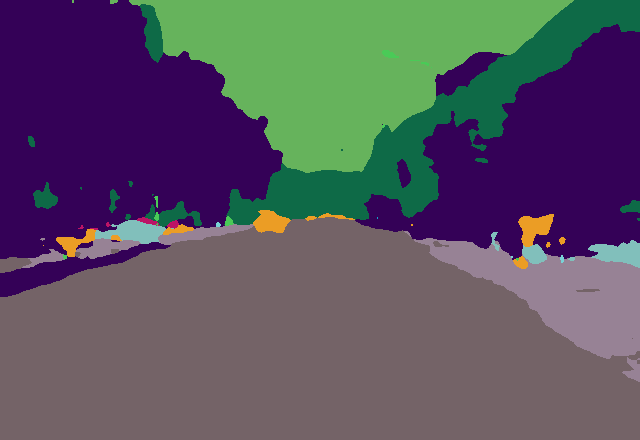}
    & \includegraphics[width=0.31\linewidth]{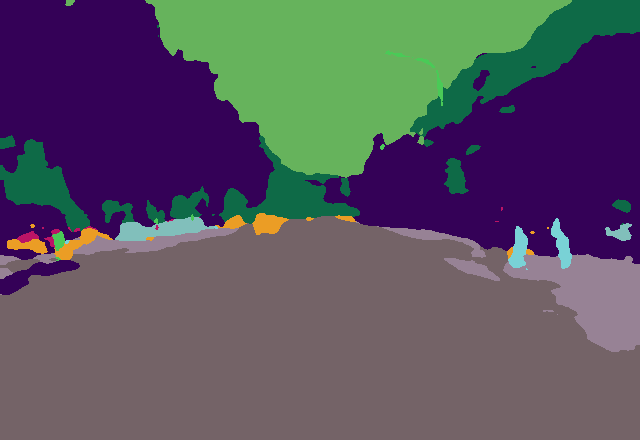}
    & \includegraphics[width=0.31\linewidth]{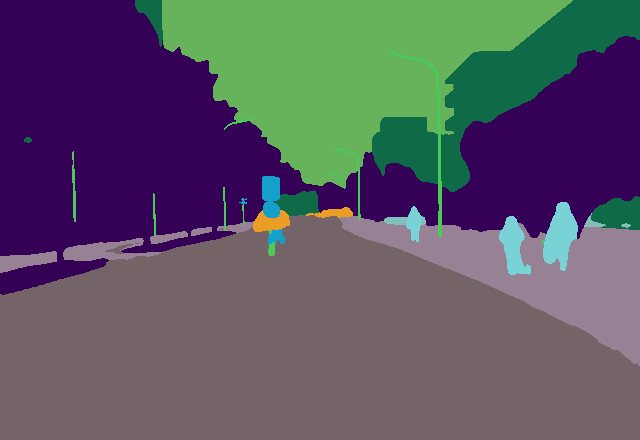}
    \\
    \includegraphics[width=0.31\linewidth]{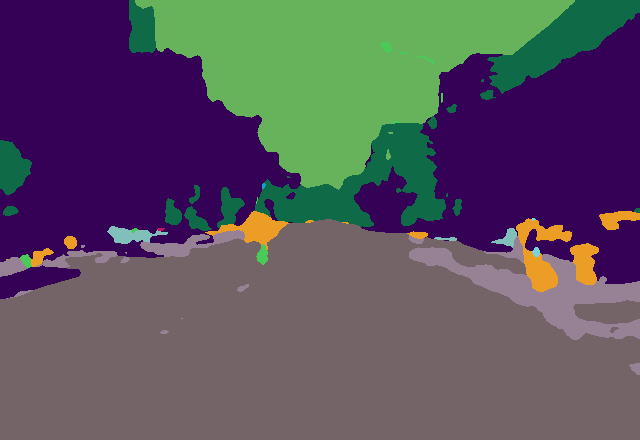}
    & \includegraphics[width=0.31\linewidth]{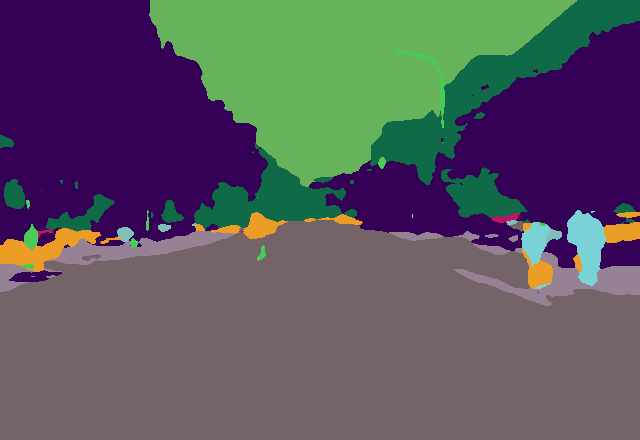}
    & \includegraphics[width=0.31\linewidth]{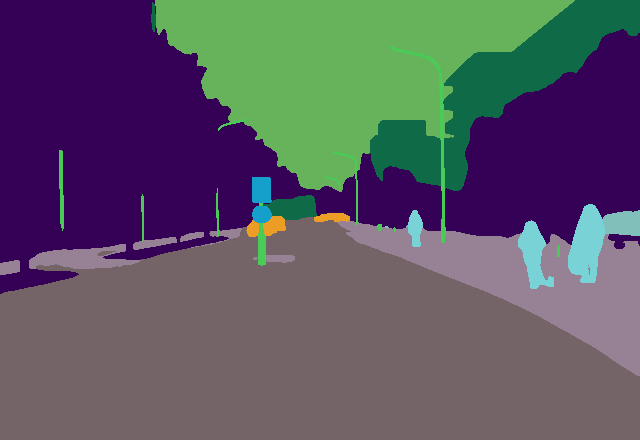}
    \\
    \includegraphics[width=0.31\linewidth]{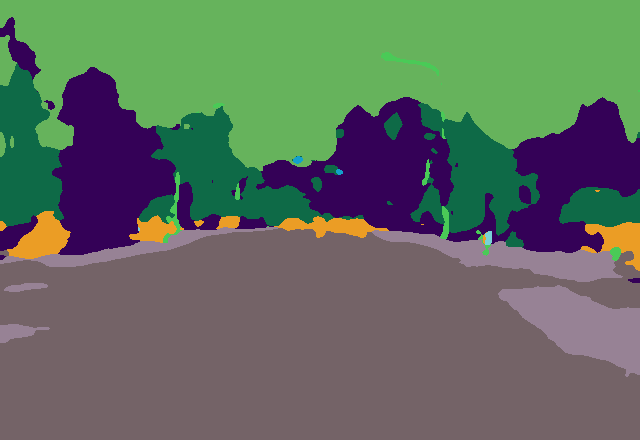}
    & \includegraphics[width=0.31\linewidth]{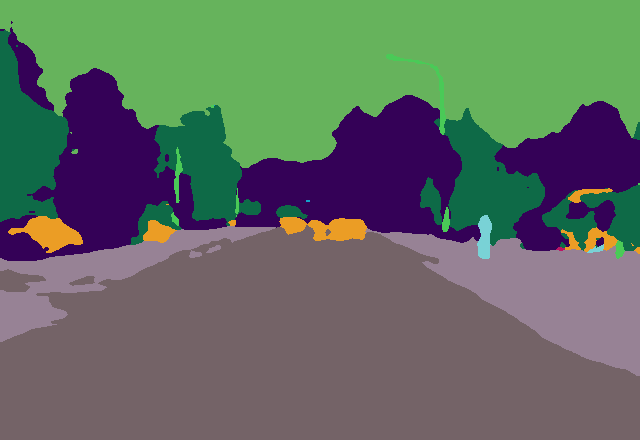}
    & \includegraphics[width=0.31\linewidth]{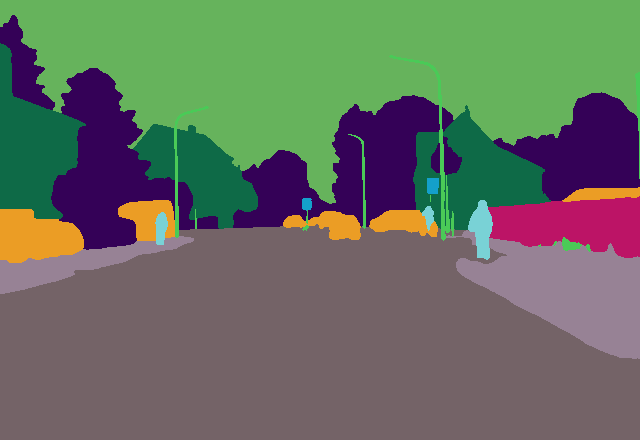}
    \\
    \includegraphics[width=0.31\linewidth]{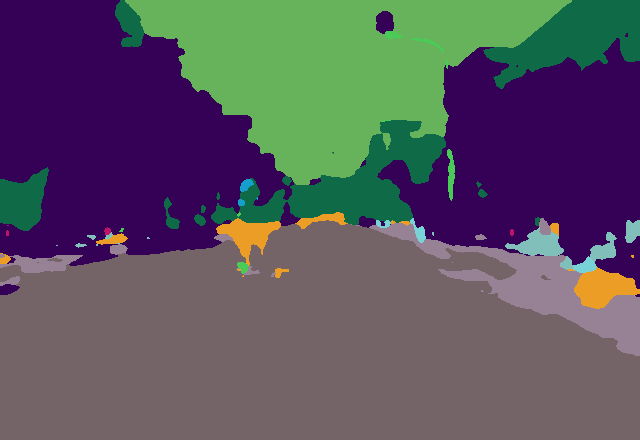}
    & \includegraphics[width=0.31\linewidth]{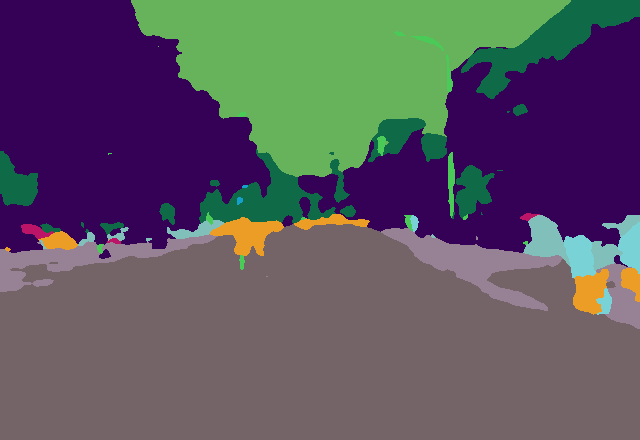}
    & \includegraphics[width=0.31\linewidth]{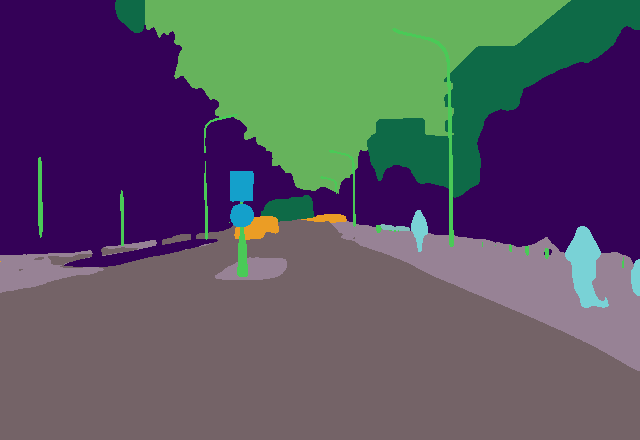}
    \\
    (a) ViT-from scratch & (b) MEM (ours)& (c) Ground truth \\
    \end{tabular}
\caption{Semantic segmentation examples: (a)~ViT-from-scratch, (b)~MEM (ours), (c) the ground truth. MEM recovers the pedestrians (the right image half, 1st - 4th row), as well as the lamp pole (the right image half, 2nd and 4th row) more reliably.}
\vspace{-0.5em}
\label{fig:ssEx2}
\end{figure*}

\begin{figure*}[h]
	\centering
    \begin{tabular}{c c c}
    \includegraphics[width=0.31\linewidth]{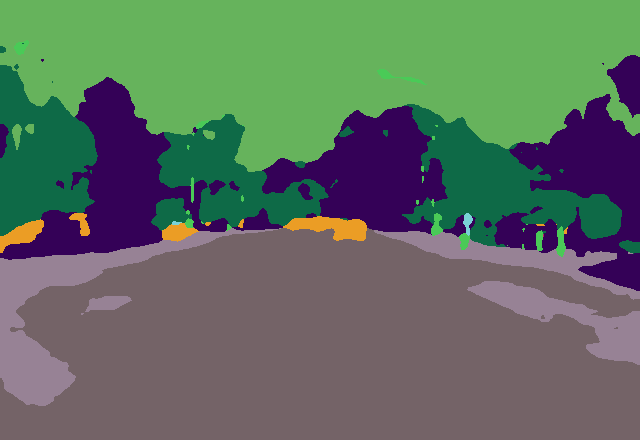}
    & \includegraphics[width=0.31\linewidth]{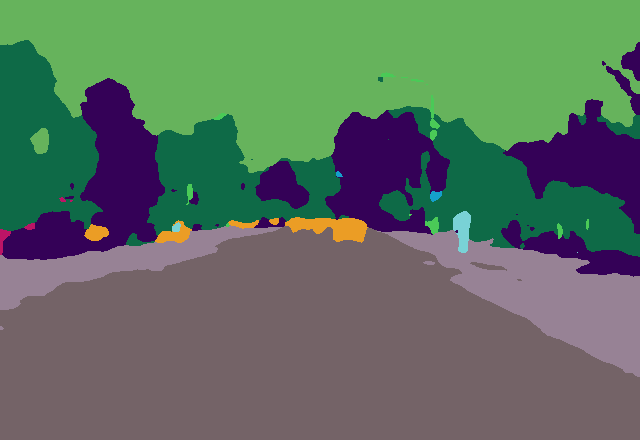}
    & \includegraphics[width=0.31\linewidth]{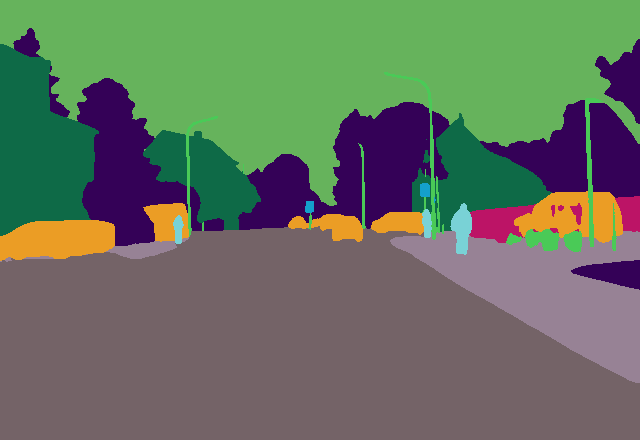}
    \\
    \includegraphics[width=0.31\linewidth]{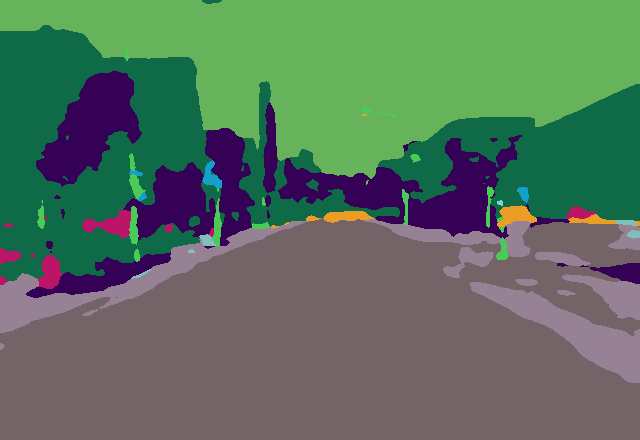}
    & \includegraphics[width=0.31\linewidth]{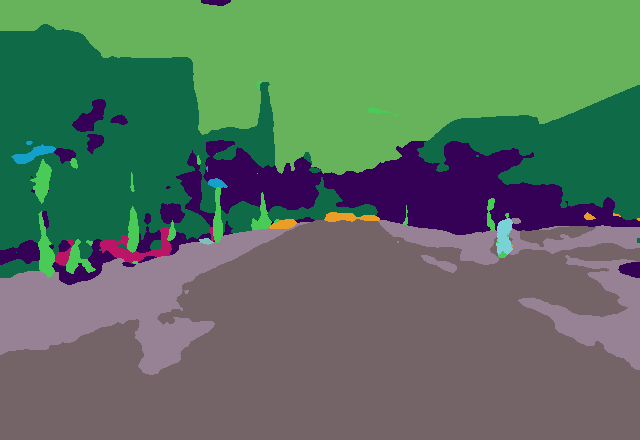}
    & \includegraphics[width=0.31\linewidth]{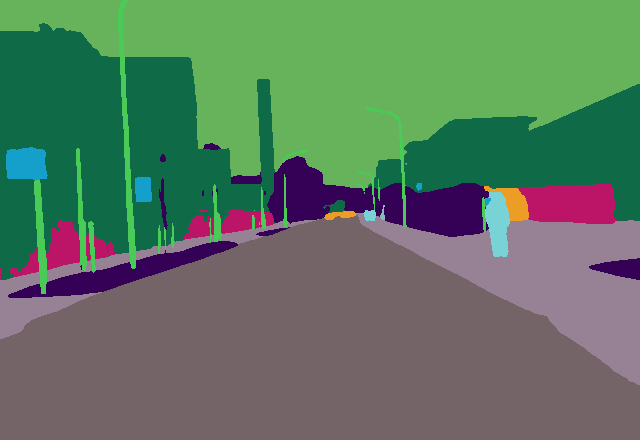}
    \\
    \includegraphics[width=0.31\linewidth]{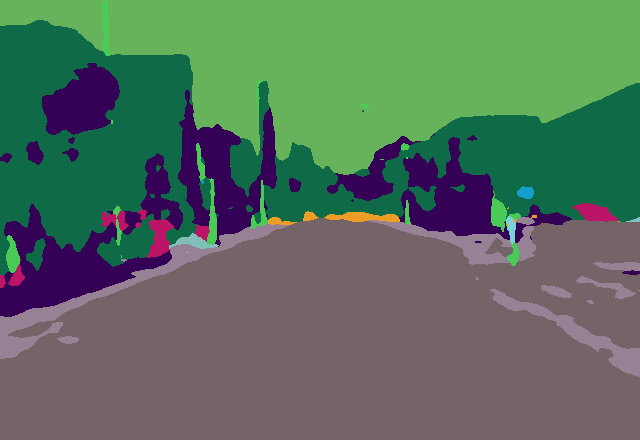}
    & \includegraphics[width=0.31\linewidth]{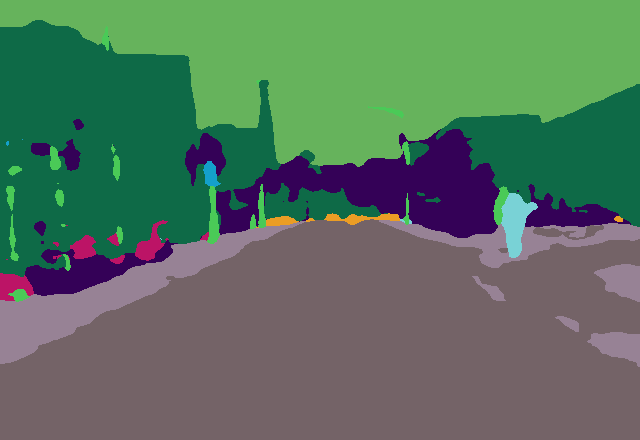}
    & \includegraphics[width=0.31\linewidth]{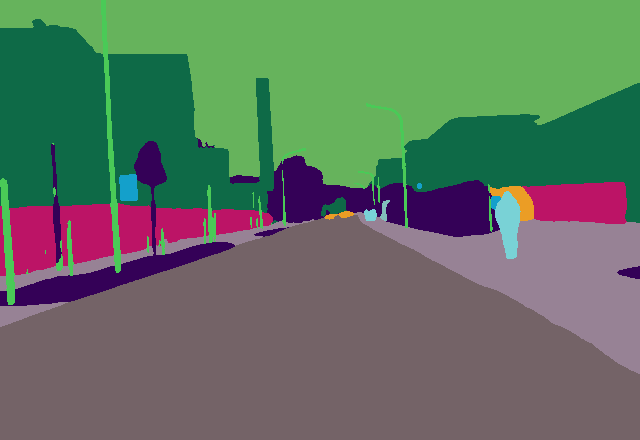}
    \\
    \includegraphics[width=0.31\linewidth]{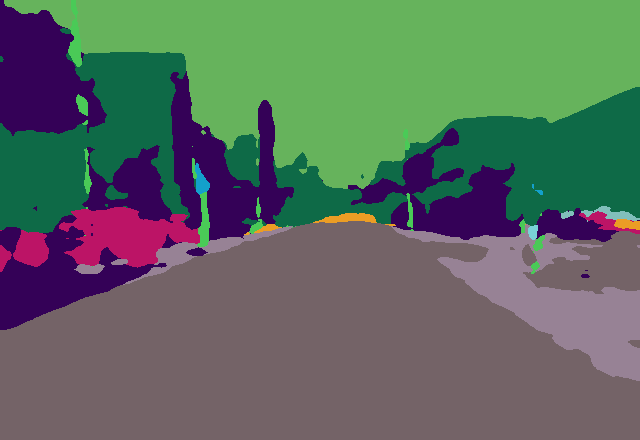}
    & \includegraphics[width=0.31\linewidth]{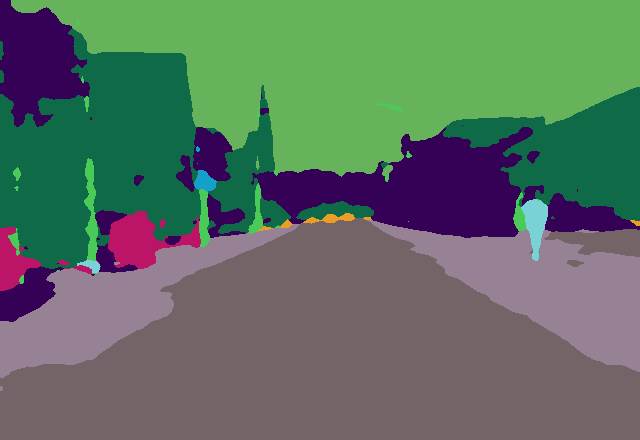}
    & \includegraphics[width=0.31\linewidth]{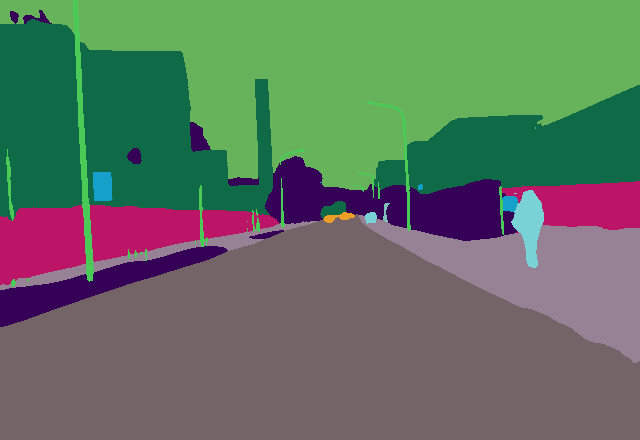}
    \\
    (a) ViT-from scratch & (b) MEM (ours)& (c) Ground truth \\
    \end{tabular}
\caption{Semantic segmentation examples: (a)~ViT-from-scratch, (b)~MEM (ours), (c) ground-truth annotation map. MEM pretraining recovers the pedestrians (the right image half, 1st - 4th row), as well as the lamp pole (the left image half, 2nd - 4th row) more reliably.}
\label{fig:ssEx3}
\end{figure*}

In \cref{fig:ssEx,fig:ssEx2,fig:ssEx3}, we show additional qualitative examples for the downstream task of semantic segmentation. As discussed in \cref{sec::ss} of the main paper, MEM-pretrained models tend to exhibit improved segmentation of fine-grained scene structures, such as pedestrians and lamp poles. 

\section{Convergence Curve}
\begin{figure*}[h]
    \centering

    \definecolor{blue}{HTML}{0A47C2}
    \definecolor{red}{HTML}{CC0029}
    \definecolor{brown}{HTML}{188B65}
    
    \begin{tikzpicture}
        \begin{axis}[width=0.95\textwidth, ylabel={Top-1 accuracy, \% (test)}, xlabel={Epoch}, legend pos=south east, legend cell align={left}, ymax=89, xmax=3000, ymin=25, xmin=0, height=8cm, ytick={10,30,40,50,60,70,80,90}, minor xtick={300}, axis x line=bottom, axis y line=left, grid=both, minor x tick num=3, log ticks with fixed point, line width=1pt, every axis plot/.append style={ultra thick}, mark=none]
            \addplot [mark=none, color=blue] table [x index=0, y index=1, col sep=comma] {figs/fig-convergence/ncaltech-convergence-pt-2.csv};
            \addplot [mark=none, dashed, color=red] table [x index=0, y index=1, col sep=comma] {figs/fig-convergence/ncaltech-convergence-no-pt.csv};
            \legend{MEM (ours), ViT-from-scratch}  
        \end{axis}
    \end{tikzpicture}
    \caption{Finetuning accuracy vs.\ epochs on N-Caltech101 \cite{sironi2018hats}. With our proposed pretraining (MEM), the accuracy increases much faster. It reaches a higher final accuracy of 85.60\% compared to finetuning without pretraining (ViT-from-scratch), where the final accuracy is only 66.94\%. Both the pretraining and the finetuning tasks use the entire N-Caltech101 train dataset.}
    \label{fig:convergence}
\end{figure*}
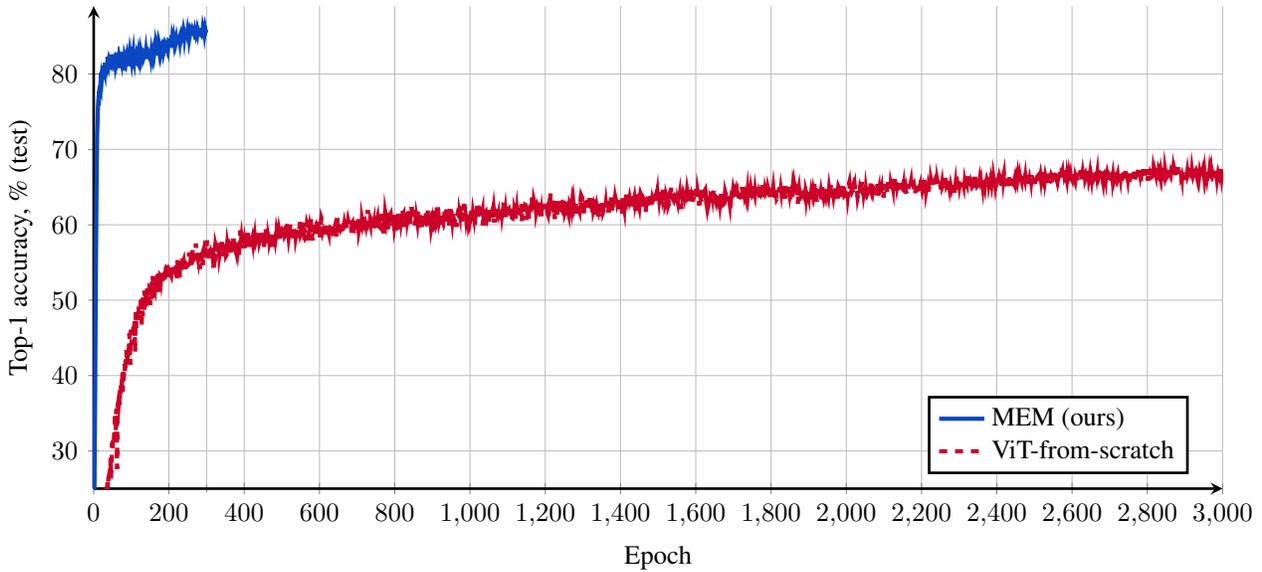

In \cref{fig:convergence}, we show that the convergence speed of finetuning is considerably higher for MEM than from ViT-from-scratch. Recall also that the top-1 accuracy of the finetuned model is substantially higher than that of the ViT-from-scratch baseline, \eg +18.66\% on N-Caltech101 in \cref{tab::ncaltech101}.

\section{Self-Supervised Baselines}
We experiment with another baseline by reconstructing frames from events using E2VID \cite{rebecq2019e2vid} to further verify our design choices.
We analyze the finetuning of both MAE and MEM on top of E2VID reconstructions in \cref{tab:e2vid-baselines}.
Both methods are strong baselines, but directly using MEM on the raw event histograms is more effective.
Moreover, the problem with a two-stage E2VID pipeline is that it uses significantly more compute and storage.
Another serious disadvantage is that the classifier uses \emph{reconstructed} images -- a lossy transformation of the raw event data, which limits the accuracy and requires domain-specific labeled training data. Analogous to \cref{tab:mae} in the main paper, we investigate two versions of MAE for completeness: E2VID+MAE-entire-hist, where the MAE loss is applied to the entire reconstructed histogram (our proposed modification for event histograms); and E2VID+MAE-only-mask, where the MAE loss is only applied to the masked patches (\cf \cite{he2022masked}). While our modified version yields a significant improvement when pretraining on raw event histograms (\cf \cref{tab:mae}), it seems to be slightly beneficial to adopt the original formulation for reconstructed frames. 

\newcolumntype{C}{>{\centering\arraybackslash}X} 
\setlength{\tabcolsep}{0em} 

\begin{table}
    \centering
    \begin{tabularx}{\linewidth}{X 
        *{4}{S[table-format=2.2,table-column-width=2.6em]}}
        \toprule
        \multirow{2}{*}{\bfseries Method} & 
        \multicolumn{2}{c}{\bfseries N-Caltech101} & 
        \multicolumn{2}{c}{\bfseries N-Cars} \\
        \cmidrule(lr){2-3} \cmidrule(lr){4-5}
        & FT & LP & FT & LP \\
        \midrule
        E2VID+MEM & 76.86 & 56.53 & \textbf{94.53} &  \textbf{90.88}  \\
        E2VID+MAE-entire-hist        & 77.45 & 58.70 &  93.09 &  88.33 \\
        E2VID+MAE-only-mask    & \textbf{78.56} & \textbf{60.22}  & 94.39 & 89.51 \\
        \midrule
        MEM (ours) & \textbf{85.60} & \textbf{71.20} & \textbf{98.55} & \textbf{97.58} \\
        \bottomrule
    \end{tabularx}
    \caption{E2VID baselines. Top-1 accuracy for finetuning (FT) and linear probing (LP) on N-Caltech101 and N-Cars.
    }
    \label{tab:e2vid-baselines}
    \vspace{-0.25cm}
\end{table}

\section{Linear probing (LP)}
We also evaluate linear probing results with MEM pretraining.
The conclusions from the main text still hold for LP: MEM-pretrained models achieve higher top-1 accuracy than ViT-from-scratch.
We report top-1 LP accuracy on N-Caltech101 and N-Cars in \cref{tab:linear-probing}.
Compared to finetuning, LP benefits more from longer pretraining (\cf \cite[Fig.\ 7]{he2022masked}).
We trained our model for 75 epochs on N-ImageNet, which is sufficient for standard finetuning, but proves to be too short for LP.
Additionally, LP requires careful hyperparameter tuning (\cf \cite[A.1 and Tab.\ 10]{he2022masked}).
While we used here the same hyperparameters for MEM as in finetuning, we could substantially improve upon the initial LP results by using new hyperparameters and removing regularization, following the insights of MAEs~\cite{he2022masked}.

\begin{table}
    \centering
    \begin{tabularx}{\linewidth}{
    X *{2}{S[table-format=2.2,table-column-width=6em]}}
        \toprule
        {\bfseries Method}  & {\bfseries N-Caltech101} & {\bfseries N-Cars} \\
        \midrule
        MEM  + LP               & \textbf{71.20} & \textbf{97.58} \\
        MEM-NImNet + LP        & {68.19} &  {89.82} \\
        Random Init + LP       & {25.34}  & {75.94} \\
        \midrule
        ViT-from-scratch   &  66.94 & 92.71 \\
        \bottomrule
    \end{tabularx}
    \caption{Top-1 accuracy of linear probing (LP) for MEM pretraining on N-Caltech101 and N-Cars. Random Init + LP is the LP accuracy of a randomly initialized ViT.
    }
    \label{tab:linear-probing}
    \vspace{-0.1cm}
\end{table}

\section{Ablation of Input Representation}
\label{supp-sec::input_representation_supp}
Event histograms already encode temporal information implicitly since they consider up to $N_\text{max}$ events per histogram and accumulate the polarities into separate channels.
As discussed in \cref{sec::input_representation}, we did not observe a significant benefit by explicitly including temporal information in the input, such as the event timestamps.
In fact, as \cref{tab::ablationInput} shows, including this information into the event histograms as a third channel leads to slightly worse top-1 classification accuracy -- compare lines \textit{(i)} and \textit{(ii)}. 

To further investigate the importance of the temporal dimension for classification, we employ an 8-channel input histogram as the input. In contrast to our default 2-channel histograms, which collapse all recent events into two channels, this 8-channel histogram distributes the stream of $N_\text{max}$ events into four equally spaced chunks of time and computes the histogram per chunk (similar to voxel grids \cite{rebecq2019e2vid}). Compared to simply using the latest timestamp in the third channel, this representation encodes the temporal information in more fine-grained manner. However, it does not yield a consistent advantage over our 2-channel baseline, as shown in lines \textit{(iii)} and \textit{(iv)} of \cref{tab::ablation}.
While the 8-channel representation somewhat improves the top-1 accuracy on N-Cars, the result is considerably worse on N-Caltech101.

The marginal benefit of the more explicit temporal encoding, as demonstrated in these experiments, has an intuitive explanation if we consider the underlying task -- object classification.
Semantic meaning is easily accessible from the \emph{spatial} context, such as object shape, rather than the temporal distribution of brightness changes, which the more explicit temporal representations provide.

\begin{table}
\centering
\begin{tabularx}{\linewidth}{X *{2}{S[table-format=2.2,table-column-width=5.5em]}}
\toprule
{\bfseries Ablation} & {\bfseries N-Caltech101}  & {\bfseries N-Cars}  \\ 
\midrule

(i)~MEM & \textbf{85.60} & \textbf{98.55} \\
(ii)~w/ timestamps (3rd chan.) & 84.90 & 98.10 \\

\midrule

\textit{with 33\% pretrain steps: } \\
(iii)~MEM (2 channels) &  \textbf{81.17} & 95.16 \\
(iv)~8-channels  &  79.70  & \textbf{95.84}  \\
\bottomrule
\end{tabularx}

\caption{A study of alternative input representations on N-Caltech101~\cite{orchard2015NCaltech} and N-Cars~\cite{sironi2018hats}. As the baseline, here we use 2-channel histograms (see \cref{sec::input_representation}) of size $224 \times 224$, patch size 16, masking ratio 50\%, RandAugment \cite{cubuk2020randaugment} and gradient clipping (see \cref{tab::dVAE,tab::PT} for other hyperparameters).}
\label{tab::ablationInput}
\end{table}

\section{Reconstruction and Token Visualization}\label{sec::viz}
We visualize additional codebook vectors in \cref{fig:patches-supp}.
We observe that codebook vectors, which have a fixed index, tend to exhibit recurring shape characteristics and a consistent preference for polarity (\eg, either positive or negative or an equal share of both). The most common codebook vectors are completely blank since the event histogram is sparsely populated with event count values. Due to redundancy, we do not visualize these blank codebook vectors. Although all codebook vectors are fixed, the decoder adapts each patch to its surroundings to form a coherent image. Hence, the visualized examples of each decoded codebook vector show some appearance variation (column-wise in \cref{fig:patches-supp}). All visualized codebook vectors are rendered using the test set of N-Cars \cite{sironi2018hats}. 

Complementing \cref{fig:recons}, \cref{fig:recons-supp} illustrates the reconstruction of masked patches during pretraining on all datasets used in this work. As we discuss in \cref{seq::mae}, the MAE pretraining task can also be employed on event histograms (in contrast to the original MAE paper \cite{he2022masked}). However, it requires the loss to be formulated on the entire histogram. We found the reconstructions from eMAE, which are visualized in \cref{fig:mae_examples}, to be not as sharp as for MEM.

\input{tables/fig-patches-supp.tex}
\input{tables/fig-recons-supp.tex}

\section{Implementation Details}\label{sec::hyp}
As discussed in \cref{sec::exps}, our MEM pretraining, as well as the baselines ViT-from-scratch, ViT-1k, and ViT-21k, share the same implementation. As detailed next, we make a significant effort to ensure strong baseline performance by using best-practice training techniques.

\pseudoparagraph{ViT Architecture}
We use ViT-Base described in \cite{dosovitskiy2020image} with patch size $16 \times 16$. It consists of 12 layers and has 12 heads in the self-attention layers. The feature size is 768, maintained by MLPs with 3072 hidden units. We add a linear projection on the ViT features during pretraining, which outputs the visual tokens. We discard this linear projection during finetuning and train a new linear layer for classification. We employ relative positional encoding \cite{dosovitskiy2020image}. 

\pseudoparagraph{Hyperparameters}
We report hyperparameters for dVAE in \cref{tab::dVAE}, pretraining in \cref{tab::PT}, and finetuning in \cref{tab::FT}. As discussed in \cref{ss::VAE}, gradient clipping is a vital hyperparameter. In \cref{tab::ablationGrad}, it can be seen that without gradient clipping, the accuracy on N-Caltech101 and N-Cars is worse than the baseline ViT-from-scratch. Hence, gradient clipping is essential to employ MEM pretraining on sparse event histograms successfully. Note that gradient clipping is not used in the RGB setting \cite{bao2021beit,he2022masked}.

\begin{table}
\centering

\begin{tabularx}{\linewidth}{X *{2}{S[table-format=2.2,table-column-width=5.5em]}}
\toprule
{\bfseries Ablation} & {\bfseries N-Caltech101}  & {\bfseries N-Cars}  \\ 
\midrule

(i)~MEM & \textbf{85.60} & \textbf{98.55} \\
(ii)~ViT-from-scratch & 66.94 & 92.71 \\ 
(iii)~MEM w/o grad. clip. & 22.87 & 90.73 \\
\bottomrule
\end{tabularx}

\caption{Ablation of gradient clipping for dVAE and pretraining stage on N-Caltech101 and N-Cars. Gradient clip values are in \cref{tab::dVAE} and \cref{tab::PT}. Gradient clipping is essential to employ MEM pretraining on sparse event histograms. Note that gradient clipping is not employed in the RGB setting \cite{bao2021beit,he2022masked}.}
\label{tab::ablationGrad}
\vspace{-0.3cm}
\end{table}

\bgroup

\begin{table*}[t]
\centering

\begin{tabularx}{\textwidth}{X c@{\hspace{3em}}c@{\hspace{3em}}c}
\toprule
Hyperparameter  &  N-ImageNet \cite{kim2021nimnet} & N-Caltech101 \cite{orchard2015NCaltech} & N-Cars \cite{sironi2018hats} \\
\midrule

Optimizer & Adam \citesupp{kingma2014adam} & Adam \citesupp{kingma2014adam} & Adam \citesupp{kingma2014adam} \\
Optimizer momentum & $\beta_1, \beta_2 = (0.9, 0.999)$ & $\beta_1, \beta_2 = (0.9, 0.999)$ & $\beta_1, \beta_2 = (0.9, 0.999)$ \\
Learning rate &  1e-3 &  2e-4 &  2e-4 \\
Learning rate schedule &  exponential (0.99) &  exponential (0.99) & exponential (0.99)  \\
Learning rate layer decay &  0.98 & 0.98  & 0.98  \\
KL weight & 1e-10 & 1e-10 & 1e-10 \\
Batch size & 512 & 192 & 192 \\
Grad clip & 1e-2 & 1e-2 & 1e-2 \\
Epochs & 50 & 300 & 300 \\

\bottomrule
\end{tabularx}

\caption{Hyperparameters for the dVAE.}
\label{tab::dVAE}
\end{table*}

\egroup
\bgroup

\begin{table*}[t]
\centering

\begin{tabularx}{\textwidth}{X c@{\hspace{3em}}c@{\hspace{3em}}c}
\toprule
Hyperparameter  &  N-ImageNet \cite{kim2021nimnet} & N-Caltech101 \cite{orchard2015NCaltech} & N-Cars \cite{sironi2018hats} \\
\midrule

Optimizer & AdamW \citesupp{loshchilov2017decoupled} & AdamW \citesupp{loshchilov2017decoupled} & AdamW \citesupp{loshchilov2017decoupled} \\
Optimizer momentum &  $\beta_1, \beta_2 = ( 0.9, 0.95)$  & $\beta_1, \beta_2 = ( 0.9, 0.95)$ & $\beta_1, \beta_2 = ( 0.9, 0.95)$ \\
Learning rate &  1e-4 &  5e-4 &  3e-4 \\
Learning rate schedule &  cosine decay \citesupp{loshchilov2016sgdr} &  cosine decay \citesupp{loshchilov2016sgdr} & cosine decay \citesupp{loshchilov2016sgdr} \\
Warmup steps & 1000 & 1000 & 1000 \\
Weight decay &  0.05 & 0.05  & 0.05  \\
Batch size & 512 & 512 & 384 \\
Grad clip & 30 & 30 & 30 \\
Epochs & 75\textsuperscript{\textdagger} & 3000 & 1000\textsuperscript{$\ddagger$} \\

\bottomrule
\end{tabularx}

\caption{Hyperparameters for pretraining. \textsuperscript{\textdagger}Cosine scheduler set for 300 epochs, but for computational reasons, only training for 75 epochs. \textsuperscript{$\ddagger$} Cosine scheduler set for 3000 epochs, but only training for 1000 epochs.}
\label{tab::PT}
\end{table*}

\egroup
\bgroup

\begin{table*}[t]
\centering

\begin{tabularx}{\textwidth}{X c@{\hspace{3em}}c@{\hspace{3em}}c}
\toprule
Hyperparameter  &  N-ImageNet \cite{kim2021nimnet} & N-Caltech101 \cite{orchard2015NCaltech} & N-Cars \cite{sironi2018hats} \\
\midrule

Optimizer & AdamW \citesupp{loshchilov2017decoupled} & AdamW \citesupp{loshchilov2017decoupled} & AdamW \citesupp{loshchilov2017decoupled} \\
Optimizer momentum & $\beta_1, \beta_2 = ( 0.9, 0.95)$  & $\beta_1, \beta_2 = ( 0.9, 0.95)$ & $\beta_1, \beta_2 = ( 0.9, 0.95)$ \\
Learning rate &  1e-3 &  4e-3 &  5e-4 \\
Learning rate schedule &  cosine decay \citesupp{loshchilov2016sgdr} &  cosine decay \citesupp{loshchilov2016sgdr} & cosine decay \citesupp{loshchilov2016sgdr} \\
Learning rate layer decay &  0.65 & 0.65  & 0.65  \\
Warmup epochs & 20 & 20 & 20  \\
Weight decay &  0.3 & 0.05  & 0.05  \\
Drop path & 0.1 & 0.1 & 0.1 \\
Dropout & 0.0 & 0.1 & 0.1 \\
Batch size & 1024 & 1024 & 1024 \\
Epochs & 200\textsuperscript{\textdagger} & 300 & 300 \\

\bottomrule
\end{tabularx}

\caption{Hyperparameters for finetuning. \textsuperscript{\textdagger}Cosine scheduler set for 300 epochs, but for computational reasons, only finetuning for 200 epochs. We report the exponential moving average accuracy on N-Imagenet with a decay factor of 0.9999.} 
\label{tab::FT}
\end{table*}

\egroup

\subsection{Details on Event Preprocessing}
After loading all events for a given sample (\eg, accumulated over $300$ms in N-Caltech101), we slice the events in time by randomly selecting one contiguous batch comprising up to $N_\text{max} = 30,000$ events. During training, we perform \textit{(i)} a random flip of all event polarities (with probability $p=0.5$); \textit{(ii)} a random horizontal flip (with probability $p=0.5$); and \textit{(iii)} random shifts (per event) by $\Delta x$ and $\Delta y$ of $x$-coordinates and $y$-coordinates using uniform sampling, \ie $\Delta x \sim \mathcal{U}(-15, 15)$ and $\Delta y \sim \mathcal{U}(-15, 15)$. 

We accumulate the augmented events into a two-channel histogram. For N-Caltech101 and N-Cars, we resize the histograms to spatial resolution $224 \times 224$. For N-ImageNet, we resize the histogram to $256 \times 341$ and randomly crop the image to $224 \times 224$. 
Next, we remove ``hot pixels'', a variant of noise specific to event cameras, which manifests as a continuously triggering event \citesupp{hu2021v2e}. We define a pixel as a ``hot pixel'' if its event count is ten standard deviations above the mean value in the event batch. We normalize the histogram values to $\left[ 0, 1 \right]$. Lastly, during training, we perform RandAugment \cite{cubuk2020randaugment} with two operations and a magnitude of 20. The three stages of MEM (dVAE, pretraining, and finetuning) share the same input preprocessing (\ie, the event histogram).

\subsection{Datasets}
We use the official train and test splits of N-ImageNet \cite{kim2021nimnet} and N-Cars \cite{sironi2018hats} and  DSEC-Semantic \cite{Gehrig21ral}. For N-Caltech101 \cite{orchard2015NCaltech}, we randomly split the data into 80\% training data and 20\% test data.
We ran 5-fold cross-validation to confirm that all random splits yield approximately the same result on N-Caltech101.
The top-1 accuracy on the test sets in these experiments were 84.6\%, 84.7\%, 85.3\%, 85.6\%  (reported in the main text), and 85.8\%. 

\input{tables/fig-mae-examples}

\vfill\null

{\small
\bibliographystylesupp{ieee_fullname}
\bibliographysupp{supp_egbib}
}

\end{document}